\documentclass{article}

     \PassOptionsToPackage{numbers, compress}{natbib}

    \usepackage[preprint]{neurips_2025}

\usepackage[utf8]{inputenc} 
\usepackage[T1]{fontenc}    
\usepackage{hyperref}       
\usepackage{url}            
\usepackage{booktabs}       
\usepackage{amsfonts}       
\usepackage{nicefrac}       
\usepackage{microtype}      
\usepackage{xcolor}         
\usepackage{microtype}
\usepackage{graphicx}
\usepackage{subfigure}
\usepackage{booktabs} 
\usepackage{wrapfig}

\usepackage{amsmath}
\usepackage{amssymb}
\usepackage{mathtools}
\usepackage{amsthm}
\usepackage{enumitem}
\usepackage{svg}
\usepackage{subcaption}

\usepackage[capitalize,noabbrev]{cleveref}

\usepackage{multirow}
\usepackage{multicol}

\usepackage[textsize=tiny]{todonotes}
\usepackage[most]{tcolorbox}

\tcbset{
  myepibox/.style={
    colback=yellow!10,       
    colframe=orange!80!black,
    arc=4mm,                 
    boxrule=1pt,             
    left=4pt,
    right=4pt,
    top=3pt,
    bottom=2pt,
    breakable,               
  }
}

\title{Epistemic Artificial Intelligence is Essential for Machine Learning Models to 
{Truly}
`Know When They Do Not Know'}

\author{\vspace{7pt}\textbf{Shireen Kudukkil Manchingal}\textsuperscript{$1$}\thanks{Corresponding author: \textit{smanchingal@brookes.ac.uk}}\ \ \ \ \ \ \ \ \ \ \ \ \ 
\textbf{Andrew Bradley}\textsuperscript{$1$}\ \ \ \ \ \ \ \ \ \ \ \ \ 
\textbf{Julian F. P. Kooij}\textsuperscript{$2$} \\ 
\vspace{7pt}
\textbf{Keivan Shariatmadar}\textsuperscript{$3$,$4$}\ \ \ \ \ \ \ \ \ \ \ \ \ \ \ \
\textbf{Neil Yorke-Smith}\textsuperscript{$5$}\ \ \ \ \ \ \ \ \ \ \ \ \ \ \ \
\textbf{Fabio Cuzzolin}\textsuperscript{$1$} \\ 
\textsuperscript{$1$}School of Engineering, Computing and Mathematics, Oxford Brookes University, UK\\
\textsuperscript{$2$}Cognitive Robotics, TU Delft, Netherlands \ \
\textsuperscript{$3$}LMSD, Mechanical Engineering, KU Leuven \\
\textsuperscript{$4$}Flanders Make@KU Leuven \ \
\textsuperscript{$5$} STAR Lab, TU Delft, Netherlands \\
\texttt{\{smanchingal, abradley, fabio.cuzzolin\}@brookes.ac.uk} \\
\texttt{\{J.F.P.Kooij, n.yorke-smith\}@tudelft.nl} \\
\texttt{\{keivan.shariatmadar\}@kuleuven.be}\\
}

\begin{document}

\maketitle

\vspace{-5pt}
\begin{abstract}
Despite AI's impressive achievements, including recent advances in generative and large language models, there remains a significant gap in the ability of AI systems to handle uncertainty and generalize beyond their training data. 
AI models 
consistently 
fail to make robust enough predictions when facing unfamiliar or adversarial data. 
Traditional machine learning approaches struggle to address this issue, 
due to an overemphasis on data fitting, 
while current uncertainty quantification approaches suffer from serious limitations.
This position paper posits a paradigm shift towards \emph{epistemic artificial intelligence}, emphasizing the need for models to learn from what they know while at the same time acknowledging their ignorance,
using the mathematics of \emph{second-order uncertainty} measures. This approach, which 
leverages the expressive power of such measures to efficiently manage
uncertainty, offers an effective way to improve the resilience and robustness of AI systems, allowing them to better handle unpredictable real-world environments.
\end{abstract}

\vspace{-2mm}
\section{Introduction}
\label{sec:intro}
\vspace{-5pt}

The success of artificial intelligence, especially deep learning \citep{lecun2015deep}, is indisputable. AI systems now perform many tasks at or above human levels, with generative models advancing into creativity \citep{ramesh2022hierarchical}, large language models (LLMs) \citep{brown2020language} excelling in language manipulation, and significant advancements in multimodal AI \citep{radford2021learning}. 
However, this has also led to inflated expectations. For instance, autonomous vehicles have been touted as imminent breakthroughs for over a decade but technological challenges still remain a barrier to worldwide deployment \citep{amodei2016concrete, wang2024survey}.
While generative models like ChatGPT create remarkable outputs, there is an increasing acknowledgment of the need to reassess AI's development path fundamentally \citep{bender2021dangers}. Speculation about uncontrolled AI evolution and debates around artificial general intelligence (AGI) often overshadow pressing challenges AI must address today \citep{russell2015research}.

\vspace{-2pt}
A significant limitation of current machine learning (ML) systems is their lack of robustness. Neural networks frequently make inaccurate and overconfident predictions when faced with uncertainties, such as out-of-distribution (OoD) samples, natural fluctuations, or adversarial disruptions \citep{papernot2016limitations, hendrycks2016baseline, zhang2021evaluating}. 
These issues become safety-critical in autonomous vehicles 
due to models struggling to generalize across the diverse scenarios \citep{fursa2021worsening, bojarski2016end} the vehicle may encounter. While efforts in overfitting mitigation \citep{srivastava2014dropout, mușat2021multi} and domain adaptation \citep{ganin2015unsupervised} 
are ongoing,
these approaches are arguably insufficient to address the fundamental challenges of robustness in a meaningful manner \citep{goodfellow2014explaining, freiesleben2023beyond, braiek2025machine, zhu2023scenario}.

\vspace{-2pt}
There is a growing consensus that the
accurate estimation of \textit{uncertainty} \citep{cuzzolin2024uncertainty} is vital to improve machine learning models' reliability \citep{senge2014reliable, kendall2017uncertainties}, with key applications to safety-critical areas such as autonomous driving \citep{tang2022prediction}, medical diagnosis \citep{lambrou2010reliable}, flood risk estimation \citep{chaudhary2022flood}, and structural health monitoring \citep{vega2022variational}.
{To fully capture the uncertainty in a system or process, it is necessary to recognize two main sources: \textit{aleatoric} (predictable, irreducible) and \textit{epistemic} (unpredictable, reducible) uncertainty. 
The former
arises from randomness in the data; a simple example of this is the coin-toss, where the data generating process has a stochastic component that cannot be reduced by any additional source of information \citep{hullermeier2021aleatoric}.}
The latter,
instead, arises from a \emph{lack of knowledge} 
about the system. For example, the odds of drawing the Ace of spades at random from a deck of cards might be assumed to be 1/52. However, this is based upon a prior assumption that this is a complete, standard deck. An `unknown' deck, however, may contain duplicates or missing cards, include jokers, or comprise multiple packs. Without this prior knowledge, the underlying model inevitably carries some uncertainty, which can be reduced with each subsequent observation. Hence, an awareness of the Socratic principle, to `know that you do not know', is of paramount importance. The main source of uncertainty in AI (but also in its science and engineering applications) is indeed the lack of a sufficient amount of data to train a model, in both quantity and quality (\textit{i.e.}, data fairly describing all regions of operation, including rare events). This uncertainty is epistemic in nature \citep{cuzzolin2024epistemic}, as it concerns the model itself, and can be reduced by collecting more data or information.


\vspace{-2pt}
While most scientists would agree that this is a profound problem,
a defining issue for AI is \emph{how} uncertainty should be managed, as
existing uncertainty quantification (UQ) methods for AI have key limitations. \textit{Bayesian} models are sensitive to prior mis-specification (with the risk of biasing the whole process) and incur heavy computational overhead \citep{https://doi.org/10.48550/arxiv.2105.06868, caprio2023imprecise}, while Bayesian Model Averaging (BMA) may dilute useful predictive information \citep{hinne2020conceptual, graefe2015limitations}. \textit{Ensemble} methods are computationally demanding \citep{DBLP:journals/corr/abs-2007-06823, he2020bayesian}.
{\textit{Conformal} predictors primarily capture aleatoric uncertainty in a frequentist stance \citep{bates2023testing}.} \textit{Evidential} approaches violate asymptotic assumptions, struggle with out-of-distribution data \citep{bengs2022pitfalls, ulmer2023prior, kopetzki2021evaluating, stadler2021graph} and exhibit high inference times (\S\ref{app:comparison-experiments}).

\vspace{-2pt}
\textbf{This position paper advocates for a paradigm shift towards an Epistemic Artificial Intelligence 
emphasizing the importance of learning while acknowledging ignorance, using \emph{second-order uncertainty measures} (\S{\ref{app:uncertainty-measures}})} capable of overcoming those limitations thanks to their greater expressive power.
\textbf{Epistemic AI rests on the `paradoxical’ principle that one should first and foremost learn from (or be ready for) the data it cannot see.} 
Prior to observing any data, the task at hand is thus completely unknown (albeit prior knowledge can be utilized to formalize the task and set a model space of solutions). The (limited) available evidence should only be used to temper our ignorance, to avoid `catastrophically forgetting’
how much we ignore about the problem.

\vspace{-2pt}
\textbf{Epistemic AI is supported by both theoretical arguments and strong empirical evidence} (Sec. \ref{sec:epi}).
\textbf{Firstly}, the use of {second-order uncertainty measures allows Epistemic AI to explicitly represent model ignorance and properly account for uncertainty due to lack of knowledge without biasing the learning process, 
unlike traditional approaches (Sec. \ref{sec:why-epi}).
\textbf{Secondly}, 
evidence is recently mounting that Epistemic AI can predict uncertainty more accurately, at lower inference times (\S{\ref{app:comparison-experiments}}), and more broadly outperform other UQ methods in terms of accuracy, robustness and calibration (Sec. \ref{sec:empirical}). 
As a result, Epistemic AI is capable of reducing the likelihood of AI systems being `surprised' by unexpected data or incapable to respond to unforeseen situations. This has enormous importance for mission-critical areas such as autonomous vehicles, or climate change and pandemic prediction, where long-term uncertainty is paramount as predictions concern the distant future and data is extremely scarce.
Large language models learning `epistemically’ from data would be less likely to commit to false statements. Bias issues could be significantly mitigated, as epistemic models would not simply mimic the training data but account for possible future data. 
This paper presents arguments in support of Epistemic AI, discusses its potential and future challenges, while acknowledging alternative views.

\vspace{-2pt}
\textbf{Paper structure.} 
Sec. \ref{sec:whyuncertainty} shows how estimating uncertainty aids robustness and adaptation. Sec. \ref{sec:uncertainty-models} reviews other models and perspectives. Sec. \ref{sec:epi} introduces Epistemic AI, its theoretical (\ref{sec:why-epi}) and  empirical (\ref{sec:empirical}) support. Sec. \ref{sec:generative-ai} explores its potential future role in generative AI. Secs. \ref{sec:challenges}–\ref{sec:conclusion} cover challenges, exciting opportunities in science and conclusions. Appendices \S\ref{app:uncertainty-measures} and \S\ref{app:uncertainty-models} recall second-order measures and models. \S\ref{app:related} further details related work; \S\ref{app:comparison-experiments} provides additional results.




\vspace{-8pt}
\section{Why Uncertainty Quantification Matters}
\label{sec:whyuncertainty}
\vspace{-4pt}

\textbf{Adversarial Robustness.}
Traditional neural networks often suffer from overconfidence (softmax outputs reflect relative confidence, not true uncertainty) leading to high-confidence errors on out-of-distribution (OoD) or adversarially perturbed inputs
\citep{guo2017calibration, hendrycks2016baseline}. Fig. \ref{fig:imagenet-a} compares a standard ResNet50 (\textit{Traditional}) and an uncertainty-aware ResNet50 (\textit{Epistemic}) \citep{manchingal2025randomsetneuralnetworksrsnn} on ImageNet-A \citep{hendrycks2021natural}, an adversarially filtered dataset exposing model overconfidence. When both models, trained on ImageNet, are tested on ImageNet-A, the Traditional model remains highly confident in its misclassifications, whereas the Epistemic model assigns lower confidence to misclassifications, avoiding overconfidence.

\vspace{-2pt}
\textbf{Robustness and Domain Adaptation.} Domain adaptation methods such as minimax learning \citep{azar2017minimax}, counterfactual error bounding \citep{swaminathan2015self}, and custom loss functions use adversarial feature alignment for unsupervised adaptation \citep{awais2021adversarial, alijani2024vision}. Reinforcement learning robustness employs adversarial strategies \citep{pinto2017robust} and Bayesian Bellman formulations \citep{derman2020bayesian}. Out-of-distribution (OoD) and domain-generalization research leverages kernel methods \citep{blanchard2011generalizing, deshmukh2019generalization, hu2020domain, 10.5555/3042817.3042820} and H-divergence–based adversarial learning \citep{albuquerque2019generalizing}, yet all falter under significant train–test shifts \citep{rosenfeld2022online}, where uncertainty management can improve adaptation and OoD detection \citep{gulrajani2020search, koh2021wilds, singh2024domain}.

\begin{wrapfigure}{r}{0.5\textwidth}
\vspace{-32pt}
    \hspace{-24pt}
    \includegraphics[width=0.56\textwidth]{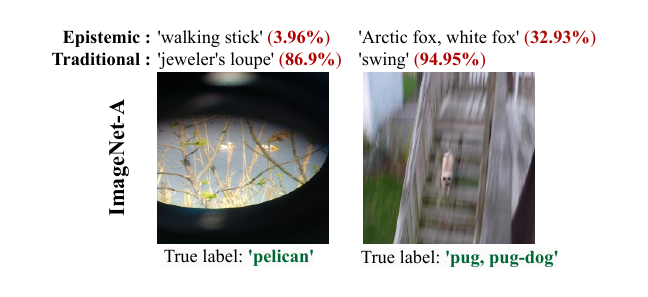}
    \vspace{-26pt}
    \caption{Confidence scores of uncertainty-aware (\textit{Epistemic}) and  (\textit{Traditional}) model on ImageNet-A (adversarial).
    Unlike the epistemic model, traditional model is overconfident in misclassifications.
    }
      \label{fig:imagenet-a}
      \vspace{-18pt}
\end{wrapfigure}
\vspace{-2pt}
\textbf{Calibration.} Neural networks are typically uncalibrated, \textit{i.e.}, predicted confidence rarely equals accuracy \citep{guo2017calibration}. Post-hoc techniques such as histogram and Bayesian binning, Platt scaling \citep{ojeda2023calibrating,platt1999probabilistic}, and regression extensions \citep{kuleshov2018accurate} improve this. Expected Calibration Error (ECE) \citep{naeini2015obtaining}, Adaptive CE \citep{nixon2019measuring}, and loss-based adjustments \citep{mukhoti2020calibrating,luo2022local,tao2023dual} refine calibration but still overlook deeper uncertainty representation.

\vspace{-2pt}
\textbf{Sequential decision‐making} must also model the propagation of uncertainty, especially in safety-critical domains such as autonomous driving \citep{teeti2023temporal}, where unmodeled perception or state uncertainty can cause compound errors and unsafe actions \citep{schwarting2018planning}. Effective quantification of {epistemic} uncertainty enables the system to detect unreliable predictions and act with `human-like' cautiousness \citep{kendall2017uncertainties,depeweg2018decomposition}.

\vspace{-9pt}
\section{Alternative Views}
\label{sec:uncertainty-models}
\vspace{-6pt}

\vspace{-2pt}
\textbf{Traditional and Deterministic methods.}
Traditional models make deterministic predictions and lack uncertainty modeling, assuming exact input-output relations. Deep Deterministic Uncertainty (DDU) \citep{mukhoti2023deep} estimates epistemic uncertainty via latent representation analysis or distance-sensitive functions rather than softmax probabilities \citep{alemi2018uncertainty, wu2020simple, liu2020simple, mukhoti2023deep, van2020uncertainty}. However, regularization techniques like bi-Lipschitz, commonly used in these models, do not effectively improve OoD detection or calibration \citep{postels2021practicality}. Unlike other methods, DDU captures uncertainty in the input space by detecting OoD samples rather than the prediction space. Both DDU and traditional models make point predictions (Fig. \ref{fig:uncertainty-models}).

\vspace{-2pt}
\textbf{Bayesian Methods.}
Bayesian Deep Learning (BDL) \citep{Buntine1991BayesianB, 10.1162/neco.1992.4.3.448, neal2012bayesian} models network parameters as distributions using Bayesian neural networks (BNNs) \citep{blundell2015weight, gal2016dropout, DBLP:journals/corr/abs-2007-06823}, producing predictive distributions by sampling from an approximated posterior \citep{hullermeier2021aleatoric}. 
Despite advances in training via sampling \citep{hoffman2014no, neal2011mcmc} and variational inference \citep{blundell2015weight, gal2016dropout, https://doi.org/10.48550/arxiv.1506.02557, hobbhahn2022fast, https://doi.org/10.48550/arxiv.1506.02158, sun2019functional, rudner2022tractable, https://doi.org/10.48550/arxiv.1506.02158}, and successes in real-world tasks
\citep{vega2022variational, kwon2020uncertainty},
practical challenges remain. This includes the significant computational complexity associated with training \citep{DBLP:journals/corr/abs-2007-06823} and inference (Tab. \ref{tab:uq-comparison-table}, \S{\ref{app:comparison-experiments}}), establishing appropriate prior distributions before training \citep{wenzel2020good, https://doi.org/10.48550/arxiv.2105.06868}, handling complex network architectures, and ensuring real-time applicability \citep{mukhoti2023deep}.
Furthermore, several studies have indicated that the use of single probability distributions to model epistemic lack of knowledge is, in fact, insufficient \citep{caprio2023imprecise, hullermeier2021aleatoric}. 

\begin{figure}[!t]
\hspace{-15pt}
    \includegraphics[width=1.06\textwidth]{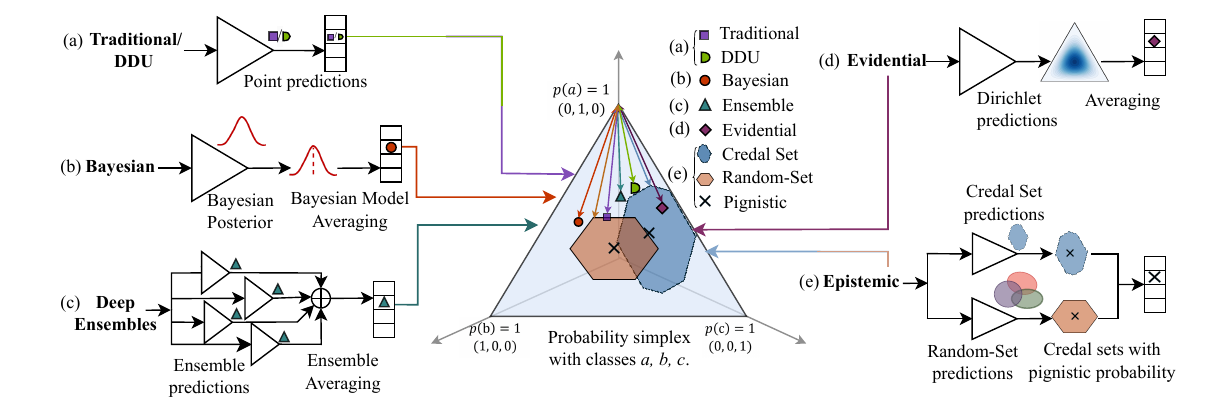}
    \vspace{-15pt}
    \caption{\textbf{Major approaches to uncertainty in AI.} Traditional networks and deterministic uncertainty models \citep{mukhoti2023deep} (a) have fixed weights and output a deterministic value or probability vector. Bayesian neural networks \citep{blundell2015weight, gal2016dropout, rudner2022tractable} (b) estimate predictive distributions by integrating over posterior parameter distributions, often approximated via Bayesian Model Averaging (BMA) \citep{hinne2020conceptual, graefe2015limitations} which averages predictions from sampled parameters. Similarly, deep ensembles \citep{lakshminarayanan2017simple} (c) average predictions from independently trained models. Evidential methods \citep{sensoy} (d) predict second-order Dirichlet parameters instead of softmax probabilities. Epistemic approaches (e) use second-order probability representations, such as interval probabilities, credal sets, or random-sets \citep{manchingal2025randomsetneuralnetworksrsnn}, with pignistic probabilities \citep{smets2005decision} derived from credal sets for comparison \citep{manchingal2025unifiedevaluationframeworkepistemic}.
}
      \label{fig:uncertainty-models}
      \vspace{-5mm}
\end{figure}

\vspace{-2pt}
\textbf{Ensemble methods} such as Deep Ensembles (DE) \citep{lakshminarayanan2017simple} and Epistemic Neural Networks (ENN) \citep{osband2024epistemic} estimate uncertainty by aggregating predictions from multiple models. DEs, in particular, have demonstrated strong performance in uncertainty estimation \citep{ovadia2019can, gustafsson2020evaluating, abe2022deep}. However, they are computationally intensive, with training and inference costs increasing linearly with the number of ensemble members, making them impractical for large models or real-time applications \citep{liu2020simple, ciosek2019conservative, he2020bayesian}.

\vspace{-2pt}
\textbf{Conformal prediction} \citep{vovk2005algorithmic} is a wrapper method applicable to any model, generating prediction sets (for accuracy guarantees) by computing empirical cumulative distributions and applying hypothesis testing to them. Several variants exist, \textit{e.g.}, conditional, full conformal prediction \citep{10.1007/3-540-36755-1_29, 10.5555/1712759.1712773, ICP, Saunders1999TransductionWC, Nouretdinov01ridgeregression, NIPS2003_10c66082, crossconformal, https://doi.org/10.48550/arxiv.1211.0025, campos2024conformal}. However, as it relies on building cumulative distribution functions of `nonconformity scores' to which it applies classical hypothesis testing \citep{bates2023testing}, conformal prediction basically models aleatoric, rather than epistemic uncertainty. Recent advances have been made towards an epistemic conformal learning, particularly under credal representations \citep{lienen2023conformal, javanmardi2024conformalized}.

\vspace{-2pt}
\textbf{Evidential Methods.}
The evidential framework \citep{155943} has been applied to neural networks \citep{ROGOVA1994777}, decision trees \citep{elouedi00decision}, K-nearest neighbours \citep{Denoeux2008classic}, and evidential deep learning classifiers for uncertainty quantification \citep{tong2021evidential}. \citet{sensoy} introduced a Dirichlet-based classifier to estimate subjective opinions. While Dirichlet-based advances exist \citep{malinin2018predictive, malinin2019reverse, malinin2019ensemble, charpentier2020posterior}, many loss functions fail to reduce epistemic uncertainty with more data, violating asymptotic assumptions \citep{bengs2022pitfalls}. Some methods rely on OoD training data, which may be unavailable or inadequate for robust detection \citep{ulmer2023prior}, and even posterior networks with normalizing flows show limitations \citep{kopetzki2021evaluating, stadler2021graph}.

\vspace{-2pt}
Some critics, including ourselves, argue that classical probability theory cannot fully address `second-level' uncertainty \citep{hullermeier2021aleatoric}, suggesting the use of more generalized frameworks (\S{\ref{app:uncertainty-measures}}), such as possibility theory \citep{Dubois90}, probability intervals \citep{halpern03book}, credal sets \citep{levi80book} or imprecise probabilities \citep{walley91book}. 
\textbf{The way epistemic uncertainty is managed and data is leveraged is, we feel, a defining issue for AI: with this position paper, we wish to contribute to this debate and indicate possible solutions through a paradigm shift which we term \emph{epistemic artificial intelligence}.}

\vspace{-2pt}
\section{Epistemic Artificial Intelligence}
\label{sec:epi}
\vspace{-2pt}

\begin{figure*}[!h]
\centering
    \includegraphics[width=0.8\linewidth]{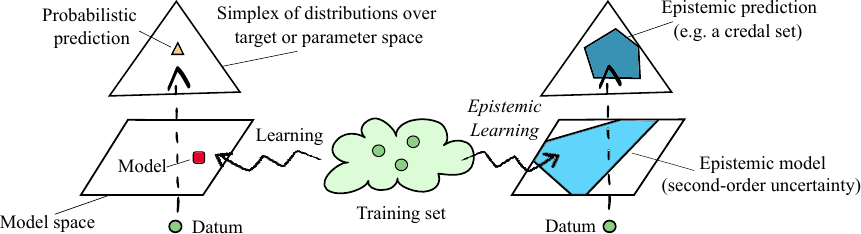}
    \caption{\textbf{Epistemic Learning.} Contrary to a traditional learning process in which a single model is learned from a training set to map new data to predictions, e.g. in the form of a probability distribution over the target space (left), epistemic learning outputs a second-order uncertainty measure (right).
}
      \label{fig:epistemicai}
      \vspace{-1.3mm}
\end{figure*}

As mentioned in Sec. \ref{sec:intro}, the core idea of Epistemic AI hinges on a paradox: the system must be designed not only to learn from the data it observes, but also to be prepared for data it has not yet encountered.
The problem can be formalized as one of learning a mapping (\textit{epistemic model}) from input data points to predictions in the form of a
second-order
uncertainty measure (\S{\ref{app:uncertainty-measures}}), either on the target space or on the parameter space of the model itself (Fig. \ref{fig:epistemicai}). Later, this prediction may be updated in the light of new data. 
When the epistemic prediction is a credal set, as in most cases \citep{manchingal2025unifiedevaluationframeworkepistemic}, a probability (`pignistic')  estimate \citep{smets2005decision} can be computed as its center of mass (Fig. \ref{fig:uncertainty-models}).

\vspace{-4pt}
\subsection{Why Epistemic AI is Essential}
\label{sec:why-epi}
\vspace{-2pt}

In addition to the limitations discussed in Sec. \ref{sec:uncertainty-models}, existing models fundamentally struggle to capture epistemic uncertainty, primarily because a \textbf{single probability distribution cannot fully express ignorance} about the data-generating process \citep{cuzzolin2020geometry}. 
Bayesian methods, in particular, though widely used, particularly falter in data-sparse or ambiguous settings because they must assign fixed belief mass even when knowledge is lacking. Uninformative priors such as Jeffreys’ \citep{jeffreys1998theory} are not invariant under reparameterization and can be improper, violating objectivity and the strong likelihood principle \citep{shafer84tech}. Moreover, priors must be specified even for systems without past data, leading to arbitrary modeling choices that can \textbf{bias the learning process for a long time} (Bernstein-von Mises theorem, \citep{kleijn2012bernstein}). Bayesian \textbf{posteriors may appear similar whether we have no knowledge or weak evidence}, conflating ignorance with imprecise belief and potentially causing misleading overconfidence. For example, as in the `unknown’ pack of cards scenario (Sec. \ref{sec:intro}), Bayesian inference treats uncertainty about the deck’s composition as a single posterior distribution, rather than explicitly quantifying our lack of knowledge. 
Model selection and prior choice lack objective criteria and \textbf{prior sensitivity worsens with scarce data} \citep{kass1996selection, berger2013statistical}. 
Further, Bayesian models \textbf{cannot naturally represent set-valued or propositional evidence}, because the additivity of probability forces allocation to individual outcomes, even when evidence supports sets of hypotheses, in opposition to random-sets which can naturally model missing data \citep{cuzzolin2020geometry}. Bayes’ rule also \textbf{assumes that new evidence is sharp and definitive}, which is unrealistic in many real-world cases.
Hierarchical Bayesian models, which place priors over priors, can model epistemic uncertainty and potentially address some of these issues, but are very computationally expensive in high-dimensional or open-world settings.

\vspace{-2pt}
Moreover, Bayesian inference tends to smooth out epistemic uncertainty by averaging over models, collapsing diverse possibilities into a single estimate and failing to distinguish knowns from unknowns \citep{hinne2020conceptual, graefe2015limitations, hullermeier2021aleatoric}. Computationally, Bayesian models also often suffer from slow convergence and large inference times (Tab. \ref{tab:uq-comparison-table}, \S{\ref{app:comparison-experiments}}), limiting their suitability for real-time safety-critical systems like autonomous vehicles \citep{DBLP:journals/corr/abs-2007-06823}.
{\textbf{We do not advocate for abandoning Bayesian approaches; rather, we argue that fully capturing epistemic uncertainty demands a generalization of Bayesian measures into broader, second‐order frameworks (\S{\ref{app:uncertainty-measures}}), calling for dedicated research and resource allocation toward these more expressive uncertainty models.}


\vspace{-2pt}

\vspace{-1pt}
\begin{tcolorbox}[myepibox]
Epistemic AI advocates for the adoption of second-order uncertainty measures, such as probability intervals, credal sets \citep{levi1980enterprise,cuzzolin2008credal} or random-sets, as they generalise classical probability using set-based representations and can richly encapsulate imprecision to model the epistemic uncertainty about an underlying, shifting data distribution, possibly the central challenge in machine learning. 
\end{tcolorbox}


\vspace{-2pt}
Indeed, most second-order measures contain classical probability as a special case \citep{cuzzolin2024uncertainty}, with random-set reasoning subsuming Bayesian reasoning as a special case \citep{shafer1981b}.

\vspace{-5pt}
\subsection{Empirical Support for Epistemic AI}
\label{sec:empirical}
\vspace{-2pt}

\begin{figure}[!ht]
    \centering
    \begin{minipage}[b]{0.49\textwidth}
        \centering
        \includegraphics[width=\textwidth]{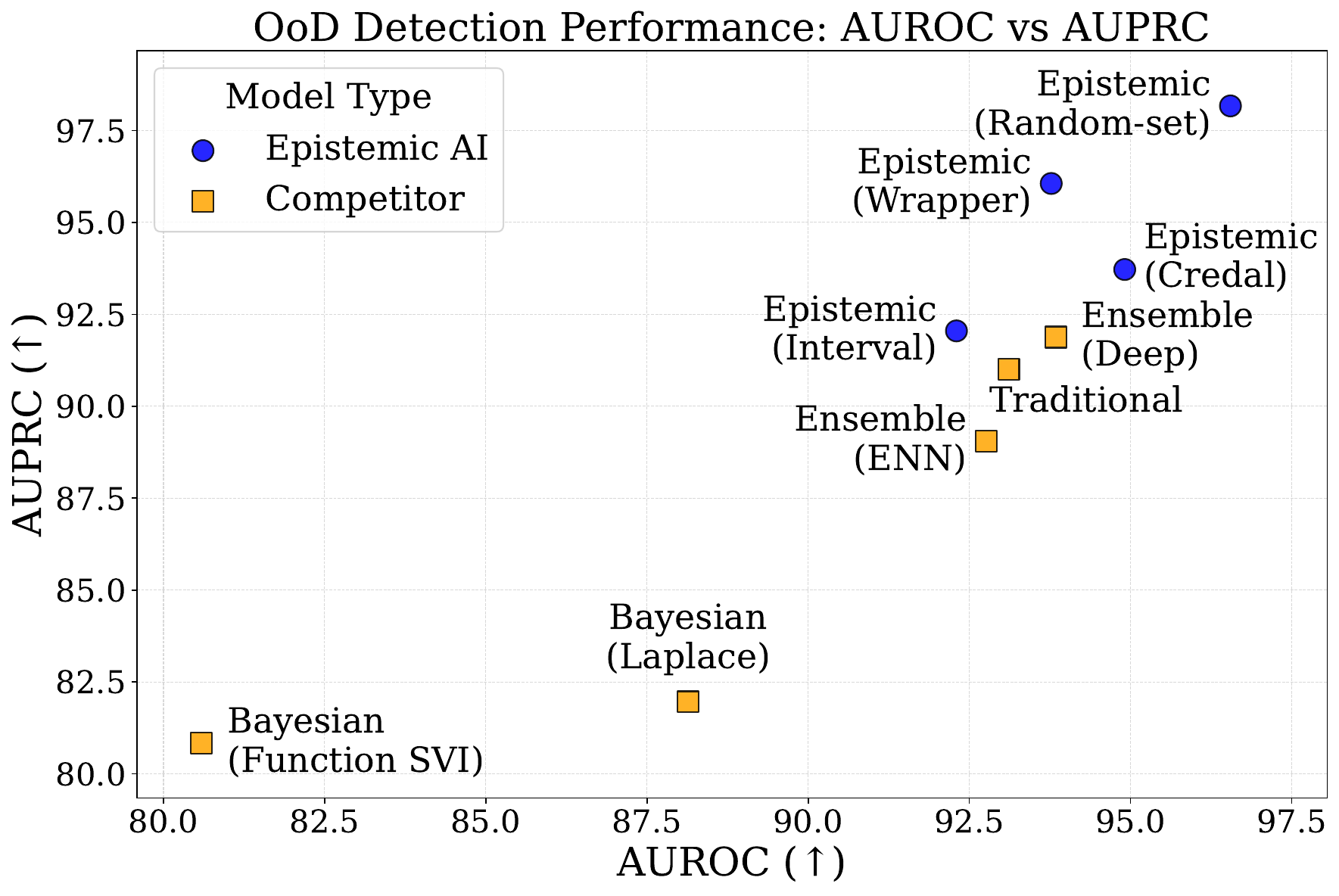}
        \vspace{-15pt}
        \caption*{(a)}
    \end{minipage}
    \hfill
    \begin{minipage}[b]{0.49\textwidth}
        \centering
        \includegraphics[width=\textwidth]{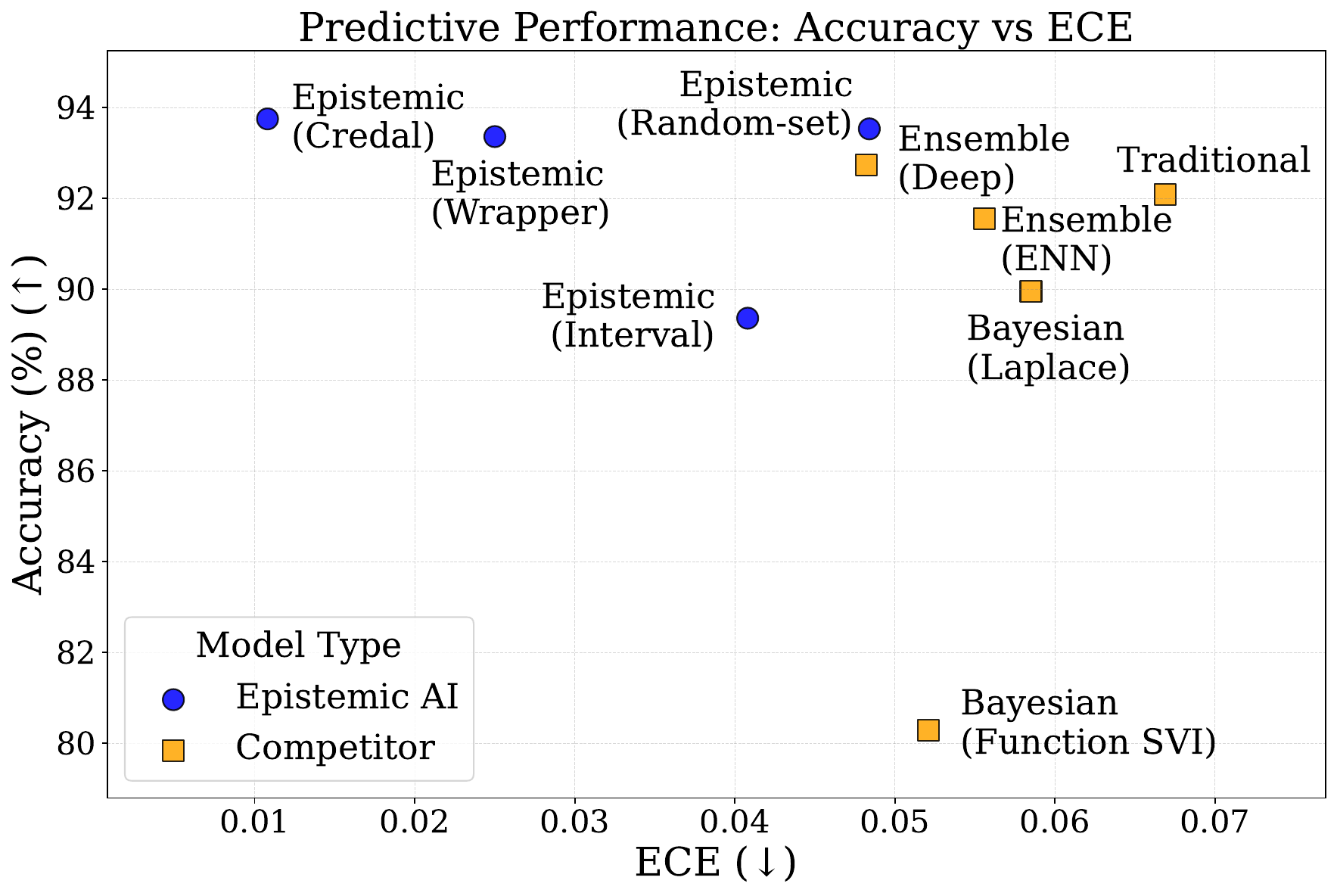}
        \vspace{-15pt}
        \caption*{(b)}
    \end{minipage}
    \vspace{-8pt}
    \caption{Comparison of \textbf{Epistemic AI models} (circles) and \textbf{competitor models} (squares) on CIFAR-10. 
    (a) \textbf{OoD detection performance (AUROC vs.\ AUPRC)}. Epistemic AI models cluster in the top-right (\textit{high separability}) while {competitor} methods show a much greater spread (\textit{lower performance}.
    (b) \textbf{Predictive performance (Accuracy vs.\ ECE).} {Epistemic AI} models cluster in the top-left (\textit{high accuracy, low calibration error}) while {competitor} methods show poorer trade-offs (\textit{weaker calibration}). Training details for all models are given in \S\ref{app:comparison-experiments}. }
    \label{fig:ood-acc-ece}
    \vspace{-8pt}
\end{figure}

Crucially, recent work on \textbf{Epistemic AI models} using second-order uncertainty measures (\textit{Epistemic: Credal} \citep{wang2024credal}, \textit{Epistemic: Wrapper} \citep{wang2024credalwrapper}, \textit{Epistemic: Random-set} \citep{manchingal2025randomsetneuralnetworksrsnn}, \textit{Epistemic: Interval} \citep{wang2025creinns})
have demonstrated superior performance over \textbf{competitor models} (\textit{Bayesian: Laplace} \citep{hobbhahn2022fast}, \textit{Bayesian: Function SVI} \citep{rudner2022tractable}, \textit{Ensemble: Deep} \citep{lakshminarayanan2017simple}, \textit{Ensemble: ENN} \citep{osband2024epistemic})
in classification tasks, based on experiments on large-scale benchmarks, including ImageNet,
in terms of accuracy, robustness, uncertainty quantification and {out-of-distribution} detection \citep{fort2021exploring},
enhancing robustness in identifying novel or anomalous inputs. 

In OoD detection, in particular, AUROC measures a model’s ability to rank OoD samples above in-distribution ones by balancing true and false positive rates across thresholds, while AUPRC focuses on the scarce OoD class by summarizing precision-recall performance. High AUROC and AUPRC, respectively, indicate strong separability and reliable detection under class imbalance, making them complementary. In Fig. \ref{fig:ood-acc-ece}(a), Epistemic AI models (circles) cluster in the top-right on CIFAR-10 \citep{recht2018cifar}, demonstrating superior, consistent OoD detection due to their ability to preserve ignorance. In contrast, competitor methods (squares) perform worse and less consistently, empirically confirming the theoretical advantage of Epistemic AI models based on second-order uncertainty measures under shifting data distributions. Moreover, Epistemic AI models also offer a better trade-off between accuracy and calibration. In Fig. \ref{fig:ood-acc-ece}(b), they dominate the top-left region, combining high accuracy with low Expected Calibration Error (ECE). More details on models, training/inference times (Tab. \ref{tab:uq-comparison-table}), and further evaluations (Fig. \ref{fig:pairwise-plots}) can be found in \S{\ref{app:comparison-experiments}}.

\vspace{-2pt}
Fig. \ref{fig:ood-acc-ece}(a) shows that second-order measures yield superior separation between in-distribution and out-of-distribution data. In Fig. \ref{fig:entropy-comparison}, the Epistemic AI model (\textit{Epistemic: Random-set}) is shown to exhibit low entropy for in-distribution and high entropy for out-of-distribution samples. This entropy gap (iD vs OoD entropy), reflected in both CIFAR-10 vs SVHN (left) and ImageNet vs ImageNet-O (right), demonstrates well-calibrated uncertainty estimates essential for reliable OoD detection and safer decision-making.

\begin{figure}[!ht]
        \centering
        \vspace{-5pt}
        \includegraphics[width=\textwidth]{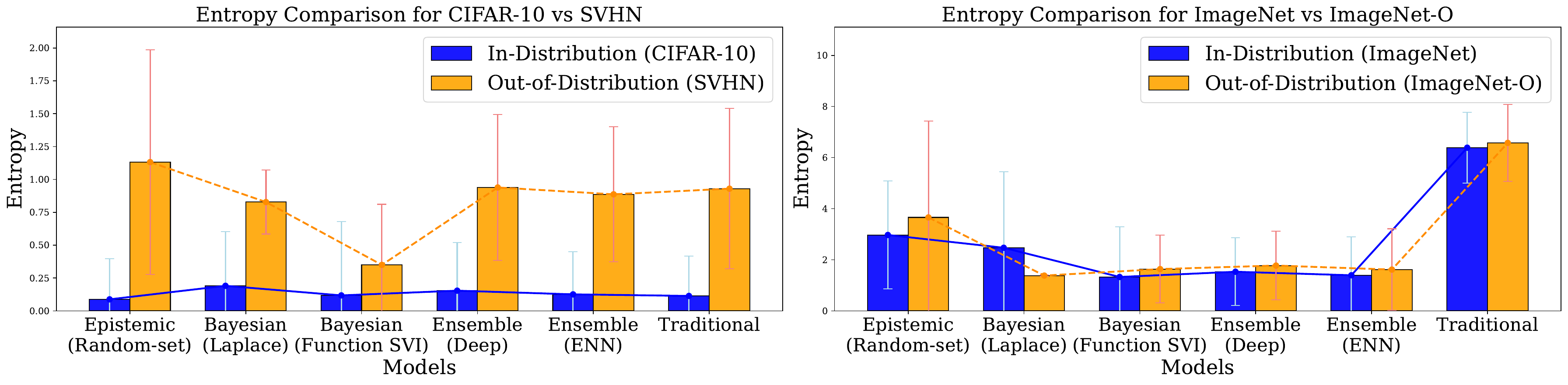}
         \vspace{-15pt}
        \caption{\textbf{Mean entropy} per model type for in-distribution (iD) and out-of-distribution (OoD) datasets, with error bars showing entropy standard deviation. The Epistemic AI model demonstrates better iD vs OoD entropy compared to other models.}
                 \vspace{-4mm}
        \label{fig:entropy-comparison}
\end{figure}

\section{Taking Epistemic AI Further}
\vspace{-4pt}

\subsection{From Target-space to Parameter-space Representations}
\label{sec:parameter-space}
\vspace{-2pt}

Epistemic uncertainty can be modelled at two levels: (i) a \textit{target} level, where the network outputs an uncertainty measure on the target space, while its parameters (weights) remain deterministic; (ii) a \textit{parameter} level, where uncertainty is modelled on the parameter space (\textit{i.e.}, weights and biases). 

The Epistemic AI models considered above are all \textbf{target-space} models.
Credal-Set Interval Neural Networks (\textit{Epistemic:Interval}) \citep{wang2025creinns}, based on Interval Neural Networks \citep{oala2021detecting}, predict probability intervals for classes. Credal Deep Ensembles (\textit{Epistemic:Credal}) \citep{wang2024credal} use ensembles of credal networks to provide upper and lower probability bounds forming a credal set; trained with a distributionally-robust optimization (DRO)-inspired loss \citep{lahoti2020fairness, nam2020learning, sagawa2019distributionally}, CreDEs outperform DEs \citep{lakshminarayanan2017simple}. Credal Wrapper \citep{wang2024credalwrapper} (\textit{Epistemic:Wrapper}) improves uncertainty estimation by ‘wrapping’ Bayesian and ensemble predictions as credal sets with upper/lower bounds per class, using the ‘intersection probability’ \citep{cuzzolin07smcb,cuzzolin09-intersection,cuzzolin2022intersection} to map a credal set to a single distribution. Random-Set Neural Networks (\textit{Epistemic:Random-set}) \citep{manchingal2025randomsetneuralnetworksrsnn} efficiently predict belief values for \emph{sets} of classes, addressing ambiguity and incomplete data, using a budgeting method to reduce complexity.

A key future research direction is thus \textbf{extending Epistemic AI from target-level to parameter-level uncertainty representations, with the aim to fully generalize Bayesian (deep) learning}. One potential approach consists of transforming Bayesian Neural Network (BNN) posteriors into random-set posteriors without retraining. Using a transform proposed by Shafer \cite{shafer1976mathematical}, any likelihood or distribution can be converted into a random-set, efficiently represented as a Dirichlet distribution over parameter intervals \cite{sultana2025epistemic}. 
Credal Bayesian Deep Learning \citep{caprio2024credalbayesian}, in opposition, introduces sets of posteriors over parameters, deriving predictive distributions at inference time that distinguish and quantify aleatoric and epistemic uncertainty, yielding either a set of outputs with guarantees or a single best prediction.
Another promising direction is to employ Smets’ Generalized Bayes Theorem (GBT) \citep{smets1993belief} (which produces belief functions \citep{cuzzolin2014belief}, \textit{i.e.}, finite random-sets, over parameters from a generalized likelihood and observations), under conditional cognitive independence (a generalization of i.i.d.), to directly learn random-set parametric representations from a training set.

\vspace{-2pt}
\textbf{Natural extensions to regression} can also be envisaged.
Credal Deep Ensembles (\textit{Epistemic:Credal}) \citep{wang2024credal} may be applied to an ensemble of Bayesian regressors, each predicting a vertex of a credal prediction, with the final credal set as their convex hull. Random-Set Neural Networks (\textit{Epistemic:Random-set}) \citep{manchingal2025randomsetneuralnetworksrsnn} may also support regression, {e.g.}, for object detection, by predicting Dirichlet distributions over Borel closed intervals \citep{strat1984continuous} for bounding box coordinates. Class labels can be modeled as sets, enabling robust uncertainty in both spatial and categorical outputs.


\vspace{-3pt}
\subsection{An Epistemic Generative AI}
\label{sec:generative-ai}

An all-important effort is ongoing to extend the epistemic paradigm to generative AI.

\textbf{Large Language Models (LLMs)} \citep{achiam2023gpt, anil2023palm, touvron2023llama} have shown strong performance in NLP tasks such as answering questions \citep{kalla2023study}, reasoning \citep{wei2022chain}, mathematical problem-solving \citep{lewkowycz2022solving}, and code generation \citep{roziere2023code}. Pre-trained on large text corpora via next-token prediction, LLMs are fine-tuned for specific applications \citep{cobbe2021training}. Despite their success, they face challenges like hallucinations \citep{maynez2020faithfulness}. Mechanisms to enhance their truthfulness (calibration) and quantify uncertainty could improve their reliability.
Bayesian approaches such as Laplace-LORA \citep{yang2023bayesian}, BLoB \citep{wang2024blob}, and Monte-Carlo Dropout (MCD) \citep{gal2016dropout}, along with techniques like Bayes by Backprop (BBB) \citep{blundell2015weight} and LORA Ensembles \citep{balabanov2024uncertainty}, have been applied to LLMs for uncertainty quantification. ENN-LLM \citep{osband2022fine} uses Epinet-inspired ensembles, while others leverage hidden states \citep{chen2024inside}, softmax entropy \citep{plaut2024softmax} or semantic entropy \citep{farquhar2024detecting}. These methods, however, often trade performance for inference efficiency.

\vspace{-2pt}
An important challenge, both in the context of LLMs and beyond, is how to elicit second-order representations from `traditional' ground truth datasets, such as question-answer pairs. \textbf{\textit{How do we teach a model that the examples it sees are only samples from an incredibly rich set of possibilities?}} Developing appropriate evaluation methods for uncertainty-aware LLMs is another challenge that needs to be addressed before such models can be effectively trained and deployed.
In the context of GenAI, \textbf{Epistemic AI can teach generative models the range of possible outputs they could produce from a limited training set}, capturing the epistemic uncertainty of the generative process itself. Our hypothesis is that modeling second-order uncertainty should enable generative models to better represent the \textit{diversity of outputs}, particularly when training data is scarce or unrepresentative. For instance, allowing LLMs to predict probabilities for \emph{sets of tokens} in a random-set framework, rather than single tokens, may allow them to capture a broader range of plausible outputs and improve overall accuracy. This could 
be especially useful for languages like Japanese or Arabic, where synonyms are prevalent and capturing a range of possible outputs is key to accurate predictions. Indeed, the random-set approach \citep{manchingal2025randomsetneuralnetworksrsnn} to classification can be directly applied to Random-Set LLMs (RS-LLMs) \citep{mubashar2025random}, where
belief functions over the vocabulary are predicted at each step instead of probability distributions, allowing language models to express ignorance. Hierarchical embedding can be used to cluster similar tokens into semantically-meaningful focal sets;
sentence uncertainty can then be calculated as the mean credal width of its tokens.
Random-set methods can also extend to generative AI via inferential models \citep{martin2013inferential, martin2015inferential}, rooted in Dempster’s belief-function theory \citep{Dempster68b} and Fisher’s fiducial inference \citep{pedersen1978fiducial}. These models can infer belief functions over neural network weights, treating generative models like GANs as auxiliary equations. Gaussian noise can be transformed into a predictive random-set, generating output variability beyond traditional methods.

\vspace{-7pt}
\section{Challenges}
\label{sec:challenges}
\vspace{-5pt}

Epistemic AI is effective and a potential key to addressing fundamental issues in machine learning. Still, challenges that are shared across uncertainty quantification may be exacerbated when using second-order uncertainty measures, owing to their higher expressiveness and complexity.

\vspace{-2pt}
\textbf{Applying second-order uncertainty measures to machine learning.}
Working with sets of distributions (\textit{e.g.}, credal or random-sets) may involve costly sampling and inference procedures, particularly for decision-making \citep{augustin2014introduction, augustin2022statistics}. Recent work has addressed this by employing set budgeting techniques to efficiently constrain the complexity of using random-sets \citep{manchingal2025randomsetneuralnetworksrsnn}. However, further research is needed to expand this to other second-order representations. 
Evaluation is an outstanding problem, as standard metrics do not apply directly to epistemic predictions. To address this, a unified framework to compare predictions across Bayesian, credal, random-set, ensemble, and evidential models was recently introduced in \citep{manchingal2025unifiedevaluationframeworkepistemic}.
Nevertheless, an accepted global metric for comparing uncertainty-aware predictions is still wanting.

\textbf{Scaling up.}
Most evidential approaches \citep{sensoy} struggle with scalability beyond medium-sized datasets. The clustering approach in the random-set approach \citep{manchingal2025randomsetneuralnetworksrsnn} has unlocked the potential of random-set representations to large datasets like ImageNet and architectures like Vision Transformers, with future extensions possibly incorporating Dirichlet mixture models \citep{yin2014dirichlet} and dynamic clustering \citep{shafeeq2012dynamic} for continual learning. A key challenge remains: \emph{can epistemic representations scale to foundation models and massive datasets?} While efficient belief function/random-set representations have been explored \citep{reineking2014belief}, further work is needed. Quantum approaches show some promise, with recent work on belief representation \citep{zhou2023bf}, combination \citep{zhou2024combining}, and integration into quantum circuits \citep{wu2024novel}.

\vspace{-2pt}
\textbf{From one-off to continual learning.}
\emph{Continual learning} is a more faithful representation of life-long real-world learning processes, especially in contexts in which models are continually updated in the light of streaming data whose distribution, however, may vary over time in unknown ways. Most research has focused on supervised learning and preventing models from `forgetting' \citep{kirkpatrick2017overcoming}, using priors, task-specific parameters, or replay buffers \citep{rolnick2019experience}. Recently, unsupervised and semi-supervised settings, such as domain-incremental learning \citep{van2019three}, have gained attention. Online learning and convex optimization \citep{dall2020optimization} offer robustness guarantees by minimizing regret. Despite recent efforts \citep{zheng2021continual, jha2024npcl}, a unified framework linking uncertainty modeling and continual learning remains an entirely open challenge, not just for Epistemic AI but for uncertainty quantification in general.

\vspace{-2pt}
\textbf{Learning and symbolic reasoning under uncertainty.}
Epistemic uncertainty can be reduced by collecting more data or incorporating prior knowledge, such as symbolic information (\textit{e.g.}, Snorkel \citep{ratner2017snorkel}), but data alone does not guarantee better performance, as seen in autonomous vehicle failures. \emph{Neurosymbolic AI} integrates symbolic reasoning with deep learning to regularize predictions and enable knowledge transfer across domains \citep{d2009neural, mao2019neuro}.
Current NeurAI frameworks enforce symbolic constraints but struggle with assessing output frequency or scaling to large knowledge bases \citep{aditya2019integrating}. Approaches like DeepProbLog \citep{manhaeve2018deepproblog} and DL2 \citep{fischer2019dl2} leverage fuzzy and probabilistic semantics but lack epistemic uncertainty modeling. Potential solutions include designing epistemic semantic losses
or using logical circuits like trigger graphs 
\citep{tsamoura2021materializing} to extend DeepProbLog-style reasoning.

\vspace{-2pt}
\textbf{Statistical guarantees.} 
Most current Epistemic AI methods do not provide statistical coverage guarantees on their predictions, albeit they can do so in combination with classical conformal learning \citep{manchingal2024randomsetneuralnetworksrsnn}. Already mentioned efforts to generalise conformal learning certainly go in this direction. Recent studies have been looking at extending the notion of confidence interval to belief functions, under the name of confidence structures \citep{denoeux2018frequency}, which generalise standard confidence distributions and generate `frequency-calibrated' belief functions. Also in the random-set setting, Inferential Models (IMs) can produce belief functions with well-defined frequentist properties \citep{martin2015inferential}. An alternative approach relies on the notion of `predictive' belief function \citep{denoeux2006constructing}, which, under repeated sampling, is less committed than the true probability distribution of interest with some prescribed probability.

\vspace{-3pt}
\section{Opportunities: Epistemic AI for Science and Engineering}
\label{sec:opportunities}

\vspace{-1.9mm}
Alongside challenges, Epistemic AI also presents a golden opportunity to enhance the AI-driven revolution in fields such as drug discovery, materials science, and astronomy. For example, DeepMind’s Alphafold \citep{jumper2021highly} revolutionized protein structure prediction, impacting {molecular biology}. Still, models like Alphafold and those used in weather forecasting often fail to model uncertainty in their predictions, which is crucial in real-world applications such as climate change, additive manufacturing or modeling of \textbf{nuclear fusion} plasma. The recently open-sourced Alphafoldv3 and {neural operator} (NO) models \citep{magnani2022approximate, garg2023vb, zou2025uncertainty},
such as those used in {nuclear fusion} and climate prediction, show promise but need better uncertainty quantification to improve accuracy and efficiency, particularly in complex systems like differential equations. 

\vspace{-2pt}
The potential of \textbf{neural operators} in Epistemic AI, as powerful surrogate models for solving PDE-governed systems across science and engineering, is significant. However, despite recent work on Bayesian \citep{magnani2022approximate} and conformal prediction (CP) \citep{gray2025guaranteed}, NOs struggle with uncertainty due to limited data or PDE misspecification. CP provides calibrated uncertainty but needs additional calibration data, which can be costly.
As neural operators learn from data a functional mapping between input and output functions (e.g., the boundary conditions and the solutions of a system of differential equations), applying epistemic learning to them involves solving the problem of \textbf{quantifying uncertainty in functional spaces}, generalising the classical neural network treatment.
Epistemic AI can help model uncertainty in the gap between low- and high-fidelity simulations, as shown in fusion plasma edge modeling \citep{faza2024interval}. 
It can also enable the robust treatment of parameter uncertainty: for instance, finite element method (FEM) simulations often estimate physical parameters within confidence bounds. 
Given the breadth of NO applications, from climate to materials science, the impact of epistemic methods is potentially profound.

\vspace{-2pt}
A paramount use case scenario is \textbf{climate change}, which is altering the weather cycle at global scale, amplifying extreme events like floods and droughts at continental scale \citep{samaniego2018anthropogenic}. Trends in the likelihood of extreme events, such as floods or droughts, are of particular interest to our society. An accurate prediction of climate change requires a correct representation of different compartments of the Earth system (\textit{e.g.} atmosphere, ocean, and land) and the interactions between them. Each of these compartments is evolving and the interaction between them is highly dynamic. 
Some limited work exists on the possible use of AI for climate change, including prediction \citep{Stein2020ArtificialIA}, mitigation \citep{Kaack2022AligningAI} and adaptation \citep{Cheong2022ArtificialIF, Chen2023ArtificialIS}. Interesting position papers and surveys on this have been published in recent years \citep{Cowls2021TheAG, Huntingford2019MachineLA}. Further, reliable long-term predictions require more than simple adaptation to a time series of data made available over time, highlighting the importance of quantifying epistemic uncertainty in the prediction of machine learning models trained on insufficient, sparse data to avoid forecasting errors and improve decision-making, with significant societal and scientific impact.

\vspace{-6pt}
\section{Conclusions}
\label{sec:conclusion}
\vspace{-4pt}

This position paper highlights the fact that \textbf{existing methods for uncertainty quantification in AI fail to efficiently model second-order uncertainty}, which is critical for epistemic uncertainty quantification and to give models the ability to `truly' know when they do not know. We argued that there is a need for much further research in this area, not only in core machine learning but also in the context of generative AI and AI for science, highlighting the need for further research and testing to further develop this promising approach. We also pointed out that significant evidence is indeed starting to support the advantage of second-order uncertainty methods in machine learning.

\vspace{-2pt}
Our \textbf{position is two-fold}: (a) We argue for the need to establish a concept we call \textbf{Epistemic AI}, according to which second-order uncertainty measures \citep{cuzzolin2021big} (\S{\ref{app:uncertainty-measures}}) are used to model epistemic uncertainty. The key argument is that ignorance is better represented through second-order uncertainty measures, which capture the inherent uncertainty about unknowns. (b) While the computational challenges of Bayesian and Ensemble models have been widely recognized, the AI community has yet to fully explore alternative models that can efficiently estimate second-order uncertainty. We also note that while there has been some progress in areas like classification and regression, significant gaps remain in more complex tasks like GenAI and AI4Science. Moreover, critical questions remain about selecting the most appropriate model for second-order uncertainty estimation and understanding the broader challenges in scaling these methods for practical applications.

\vspace{-2pt}
The recent breakthroughs in this area were made possible by realizing that it is not necessary to exploit the full expressive power of second-order uncertainty measures (\S{\ref{app:uncertainty-measures}}) to achieve significant improvements. Effective scalability can be attained by designing structures rich enough to harness the representational potential of second-order uncertainty measures while remaining computationally feasible. This can be achieved, e.g., through a suitable collection of focal sets \citep{manchingal2025randomsetneuralnetworksrsnn}, lower/upper probability structures \citep{wang2024creinns}, or a fixed budget of vertices for credal representations. Building on these results, a more principled and systematic exploration of these structures is now necessary to fully realize the vision of this paper.

\vspace{-2pt}
An exciting future research direction would be the formal definition and study of specific families of random-sets, analogous to the families of probability distributions in classical probability (Gamma, exponential, etc.), leading to more efficient and scalable computational models and driving further AI advances.
The integration of Epistemic AI with continual, neurosymbolic and neural operator learning poses a set of exciting challenges moving forward.

\section*{Acknowledgement}
This work has received funding from the European Union’s Horizon 2020 Research and Innovation program under Grant Agreement No. 964505 (E-pi). We thank the entire team of E-pi for the insightful discussions: Matthijs Spaan, Hans Hallez, David Moens, Maryam Sultana, Guopeng Li, Moritz Zanger, Pascal van der Vaart, Noah Schutte, Muhammad Mubashar, Kaizheng Wang, Adam Faza.




\bibliography{bibliography}
\bibliographystyle{apalike}

\newpage
\appendix
\onecolumn
\section{Theories of uncertainty}
\label{app:uncertainty-measures}

Uncertainty theory (UT) is an array of theories devised to encode `second-order’, `epistemic’ uncertainty, \textit{i.e.}, uncertainty about what probabilistic process actually generates the data, can provide a principled solution to this conundrum \citep{augustin2014introduction, walley91book}. This is the situation ML is in, for we usually ignore the form of the data-generating process at hand, even accepting that it should be modelled by a probability distribution. Many (but not all) uncertainty measures amount to convex sets of distributions or `credal sets’ (\textit{e.g.}, p-boxes) \citep{troffaes07}, while random-sets and belief functions directly assign probability values to sets of outcomes \citep{shafer1976mathematical}, modeling the fact that observations often come in the form of sets. The paramount principle in UT is to continually refine one’s degree of uncertainty (measured, \textit{e.g.}, by how wide a convex set of models is) in the light of new evidence. All uncertainty theories are equipped with operators (playing the role of Bayes’ rule in classical probability) allowing one to reason with such measures (\textit{e.g.} Dempster’s combination for belief functions) \citep{dempster2008upper, smets1994transferable}.

\begin{figure*}[!h]
\centering
    \includegraphics[width=0.85\textwidth]{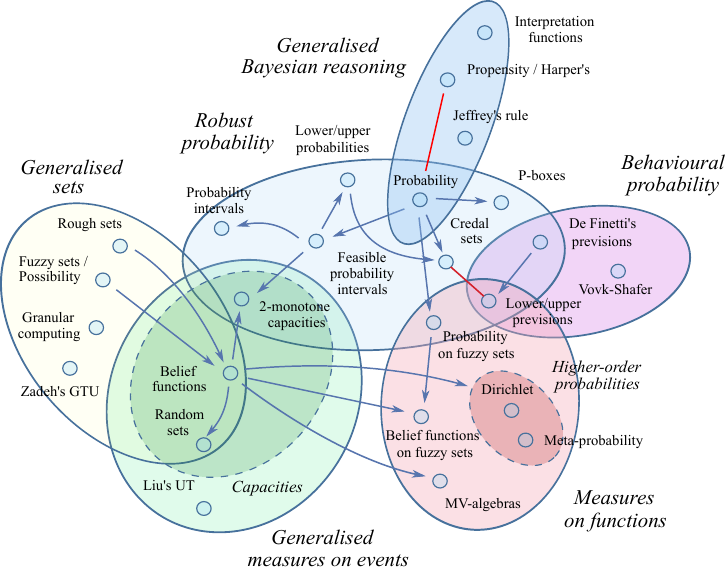}
    \caption{\textbf{Clusters of uncertainty theories.} Uncertainty theories can be arranged into various clusters based on the objects they quantify and their rationale. Arrows indicate the level of generality, with more general theories encompassing less general ones. Note that the quality and rigour of different approaches can vary significantly.
}
      \label{fig:uncertaintymeasures}
\end{figure*}

The various theories of uncertainty form `clusters’ characterised by a common rationale \citep{cuzzolin2020geometry} (see Fig. \ref{fig:uncertaintymeasures}). A first set of methods can be seen as ways of `robustifying’ classical probability: The most general such approach is Walley’s theory of imprecise probability \citep{walley91book}, a behavioural approach whose roots can be found in the ground-breaking work of de Finetti \citep{DeFinetti74}. In behavioural probability, the latter is a measure of an agent’s propensity to gamble on the uncertain outcomes. A different cluster of approaches hinges on generalising the very notion of set: these include, for instance, the theory of rough sets \citep{Pawlak1982}, possibility theory \citep{dubois2012possibility}, and Dempster-Shafer theory \citep{shafer1976mathematical}. More general still are frameworks generalising measure theory, \textit{e.g.} the theory of monotone capacities or `fuzzy measures’ \citep{grabisch2000book}. Some proposals (including Popper’s propensity) aim at generalising Bayesian reasoning \citep{popper1959propensity} in terms of either the measures used or the inference mechanisms. The theory of set-valued random variable or `random-sets’ extends the notion of set \citep{Molchanov_2017} and generalises Bayesian reasoning. Frameworks which completely replace events by scoring functions \citep{vovk2005algorithmic} in a functional space form arguably the most general class of methods. The diagram also illustrates the relationships between different theories, indicating which are more general and which are more specific. An arrow from formalism 1 to formalism 2 suggests that the former is a less general case of the latter. 

\textbf{Credal sets.} In decision theory and probabilistic reasoning, \textit{credal sets} provide a robust approach to modeling epistemic uncertainty by generalizing the traditional Bayesian framework. Unlike standard probabilistic models that assign precise probabilities to events, credal sets represent \textbf{convex sets of probability distributions}, allowing for a more flexible and cautious representation of uncertainty~\citep{walley91book}.

A \textit{credal set} is a closed and convex set of probability distributions over a finite space $\Omega$.
This formulation allows one to express \textit{imprecise probabilities}, where instead of a single probability value $P(A)$, we consider a range $[P^-(A), P^+(A)]$ that characterizes the lower and upper bounds of belief for an event $A$. This approach is particularly useful in settings where data is scarce or conflicting, making precise probability assignments unreliable~\citep{augustin2014introduction}.

Credal sets have been extensively used in \textit{robust Bayesian inference}, \textit{classification}, and \textit{decision-making under ambiguity}. In machine learning, credal classifiers~\citep{Zaffalon} extend Bayesian classifiers by considering sets of posterior probabilities rather than single estimates, improving robustness to small-sample uncertainties.

Moreover, \textit{credal networks} (generalizations of Bayesian networks) allow for imprecise conditional probability tables, leading to more cautious yet reliable inferences in high-stakes applications such as medical diagnosis and risk assessment~\citep{cozman00credal}.
By accounting for multiple possible distributions, credal sets reduce overconfidence in decision-making. Unlike Bayesian models that rely on precise priors, credal sets allow a more agnostic approach. It is particularly useful when probability estimates come from conflicting or incomplete sources. Several credal set computation techniques are discussed in \citep{manchingal2025unifiedevaluationframeworkepistemic, hullermeier2021aleatoric}.

\textbf{Challenges in credal set computations.}
Credal sets, representing convex sets of probability distributions to model uncertainty, can be handled through various computational methods. One approach involves representing a credal set by its extremal points (vertices), forming a convex polytope in the probability space. These vertices can be computed using linear programming techniques \citep{wang2024credalwrapper}, such as the simplex method, which navigates between vertices to find optimal solutions \citep{de2004inference}. Alternatively, the double description method can enumerate all vertices of a convex polytope given its defining inequalities \citep{fukuda1995double}. However, as the complexity of the network increases, the number of vertices can grow exponentially, leading to computational challenges. To address this, constraint-based representations define the credal set by a set of linear inequalities, offering computational efficiency, especially when the number of constraints is limited \citep{troffaes2014lower}. 

In the context of belief functions, credal sets can be derived through permutations of focal elements, but the combinatorial explosion necessitates optimization methods to manage computational load \citep{manchingal2025randomsetneuralnetworksrsnn}. Additionally, dual representations utilize lower and upper probabilities to perform computations without explicitly considering all extreme points \citep{wang2024credal}. The choice of method depends on the specific application and the trade-off between computational efficiency and the precision required in representing uncertainty. A study \citep{sale2023volume} found that while this volume correlates with epistemic uncertainty in binary classification, its effectiveness diminishes in multi-class classification scenarios. In contrast, more recent research \citep{javanmardi2024conformalized} indicates that the size of the credal set remains a reliable measure of epistemic uncertainty, even in multi-class settings, including complex datasets like ImageNet.

However, this paper does not simply advocate for credal sets, but for the adoption of second-order uncertainty measures. 

\vspace{-8pt}
\section{Related Epistemic AI work}
\label{app:related}
\vspace{-4pt}

Credal inference \citep{corani2012bayesian, hullermeier2021aleatoric, sale2023volume} is gradually gaining popularity as it predicts convex sets of probability distributions, known as credal sets \citep{levi1980enterprise}, providing an alternative method for efficiently quantifying epistemic uncertainty. Credal representations \citep{cuzzolin2011consistent} have been widely explored in machine learning, including the naive credal classifier \citep{corani2008learning}, credal network \citep{corani2012bayesian}, and credal random forest classification \citep{shaker2021ensemble}. 
Random-sets \citep{cuzzolin2020geometry} can naturally model missing data.
Belief function models \citep{cuzzolin2014belief,cuzzolin2010three}, in particular, have been used for ensemble classification \citep{liu2019evidence}, regression \citep{gong2017belief} or to generalise max-entropy classification \citep{cuzzolin2018generalised}, among others.

\vspace{-5pt}
\subsection{Epistemic learning theory}
\label{sec:statistical}
\vspace{-2pt}

Epistemic statistical learning theory, based on a `credal' framework \citep{caprio2024credal}, models data-generating variability via convex sets of probabilities (credal sets) inferred from finite samples. It derives bounds for finite hypothesis spaces (with or without realizability) and infinite model spaces, generalizing classical results. Data-dependent uniform PAC generalization bounds are also established using a random-set formulation \citep{dupuis2024uniform}.

\vspace{-5pt}
\subsection{Unsupervised learning}
\label{sec:unsupervised}
\vspace{-2pt}

Unsupervised clustering is central to epistemic uncertainty research. Hard methods like c-means assign objects to single clusters, while soft methods model uncertainty, including fuzzy sets \citep{3_bezdek1999fuzzy}, possibility theory \citep{31_krishnapuram1993possibilistic}, rough sets \citep{41_peters2014rough}, and evidential clustering \citep{denoeux2015ek, denoeux2004evclus}. Rough sets use approximations, while evidential clustering, based on Dempster-Shafer theory \citep{dempster2008upper, shafer1976mathematical}, represents uncertainty via mass functions, forming credal partitions. The ECM algorithm \citep{36_masson2008ecm} introduced mass-based uncertainty modeling, refined by RECM \citep{37_masson2009recm} for dissimilarity data. BCM \citep{34_liu2012belief} and CCM \citep{35_liu2015credal} addressed meta-cluster prototype issues, while BPEC \citep{49_su2018bpec}, MECM \citep{62_zhou2015median}, and EGMM \citep{jiao2022egmm} integrated evidential reasoning. EK-NN \citep{376493}, EK-NNclus \citep{denoeux2015ek}, and EVCLUS \citep{denoeux2004evclus} tackled clustering ambiguity, with NN-EVCLUS \citep{denoeux2021nn} reducing parameter dependence and enabling classification via neural networks. Key challenges include scalability, handling high-dimensional data, and ensuring robustness in uncertain environments.

\vspace{-5pt}
\subsection{Reinforcement learning}
\label{sec:reinforcement}
\vspace{-2pt}

Uncertainty quantification in reinforcement learning (RL) remains challenging, with existing methods showing practical success but lacking theoretical soundness and convergence guarantees. 
Diverse Projection Ensembles \citep{zanger2023diverse} extend distributional RL by using ensemble diversity to capture epistemic uncertainty, while still modeling aleatoric uncertainty through the distribution of returns. 
Methods like SMC-DQN \citep{van2024bayesian} combine Sequential Monte Carlo with Deep Q Networks to train model ensembles for Bayesian posterior approximation of the value function. In model-based RL, Monte Carlo Tree Search (MCTS) \citep{browne2012survey}, used in AlphaZero and MuZero, is augmented with epistemic uncertainty estimates \citep{oren2022mcts} to enhance strategic exploration. Research on Partially Observed Markov Decision Processes (POMDPs) \citep{krishnamurthy2016partially} under epistemic uncertainty includes approaches such as Bayesian POMDPs \citep{ross2007bayes} and set-valued transitions \citep{bueno2017modeling, delgado2009representing, delgado2011efficient}.
However, comprehensive extensions to emission probabilities and reward functions under various epistemic uncertainty types (intervals, credal/random-sets) are still lacking.

\section{Uncertainty estimation in uncertainty-aware models}
\label{app:uncertainty-models}

The predictions of a classifier can be plotted in the simplex (convex hull) $\mathcal{P}$ of the one-hot probability vectors assigning probability 1 to a particular class.
For instance, in a $3$-class classification scenario ($\mathbf{Y}= \{a, b, c\}$), 
the simplex would be a 2D simplex (triangle) connecting three points, each representing one of the classes, as shown in Fig. \ref{fig:uncertainty-models} (right), which 
depicts all types of model predictions considered here.

\vspace{-2pt}
\subsection{Traditional Neural Networks} 

Traditional neural networks (NNs)
predict a vector of $N$ scores, one for each class, duly \emph{calibrated} to a probability vector representing a (discrete, categorical)
probability distribution over the list of classes $\mathbf{Y}$, 
$
    \hat{p}_{NN}(y \mid \mathbf{x}, \mathbb{D}),
$
which represents the probability of observing class $y$ given the input $\mathbf{x}$ and training data $\mathbb{D}$.

\subsection{Bayesian Neural Networks} 

Bayesian Neural Networks (BNNs) \citep{lampinen2001bayesian, titterington2004bayesian, goan2020bayesian, hobbhahn2022fast} 
compute a predictive distribution $\hat{p}_{b}(y \mid \mathbf{x}, \mathbb{D})$ by integrating over a learnt posterior distribution of model parameters $\theta$ given training data $\mathbb{D}$. 
This
is often infeasible due to the complexity of the posterior, leading to the use of \emph{Bayesian Model Averaging} (BMA), which approximates the predictive distribution by averaging over predictions from multiple samples. When applied to classification, BMA yields
point-wise predictions. 


Bayesian inference integrates over the posterior distribution $p(\theta \mid \mathbb{D})$
over model parameters $\theta$ given training data $\mathbb{D}$ to compute the predictive distribution $\hat{p}_{b}(y \mid \mathbf{x}, \mathbb{D})$, reflecting updated beliefs after observing the data:
\begin{equation}
    \hat{p}_{b}(y \mid \mathbf{x}, \mathbb{D}) = \int p(y \mid \mathbf{x}, \theta) p(\theta \mid \mathbb{D}) d\theta,
\end{equation}
where $p(y \mid \mathbf{x}, \theta)$ represents the likelihood function of observing label $y$ given $x$ and $\theta$. 
To overcome the infeasibility of this integral, direct sampling from $\hat{p}_{b}(y \mid \mathbf{x}, \mathbb{D})$ using methods such as Monte-Carlo are applied to obtain a large set of sample weight vectors, $\{ \theta_k, k\}$, from the posterior distribution. These sample weight vectors are then used to compute a set of possible outputs $y_k$,
namely: 
\begin{equation}
    \hat{p}_{b}(y_k\mid \mathbf{x}, \mathbb{D}) = \frac{1}{|\Theta|} \sum_{\theta_k \in \Theta} \Phi_{\theta_k}(\mathbf{x}),
\end{equation}
where $\Theta$ is the set of sampled weights, $\Phi_{\theta_k}(\mathbf{x})$ is the prediction made by the model with weights $\theta_k$ for input $\mathbf{x}$, and $\Phi$ is the function for the model. This process is called \emph{Bayesian Model Averaging (BMA)}.
BMA may inadvertently smooth out predictive distributions, diluting the inherent uncertainty present in individual models \citep{hinne2020conceptual, graefe2015limitations} as shown in Fig. \ref{fig:BMA_vs_nonBMA}.
When applied to classification, BMA yields
point-wise predictions. For fair comparison and to overcome BMA's limitations, 
in this paper we also use sets of prediction samples obtained from the different posterior weights before averaging.

\begin{figure}[!h]
    \centering
    \begin{minipage}[t]{0.49\textwidth}
        \centering
        \includegraphics[width=\textwidth]{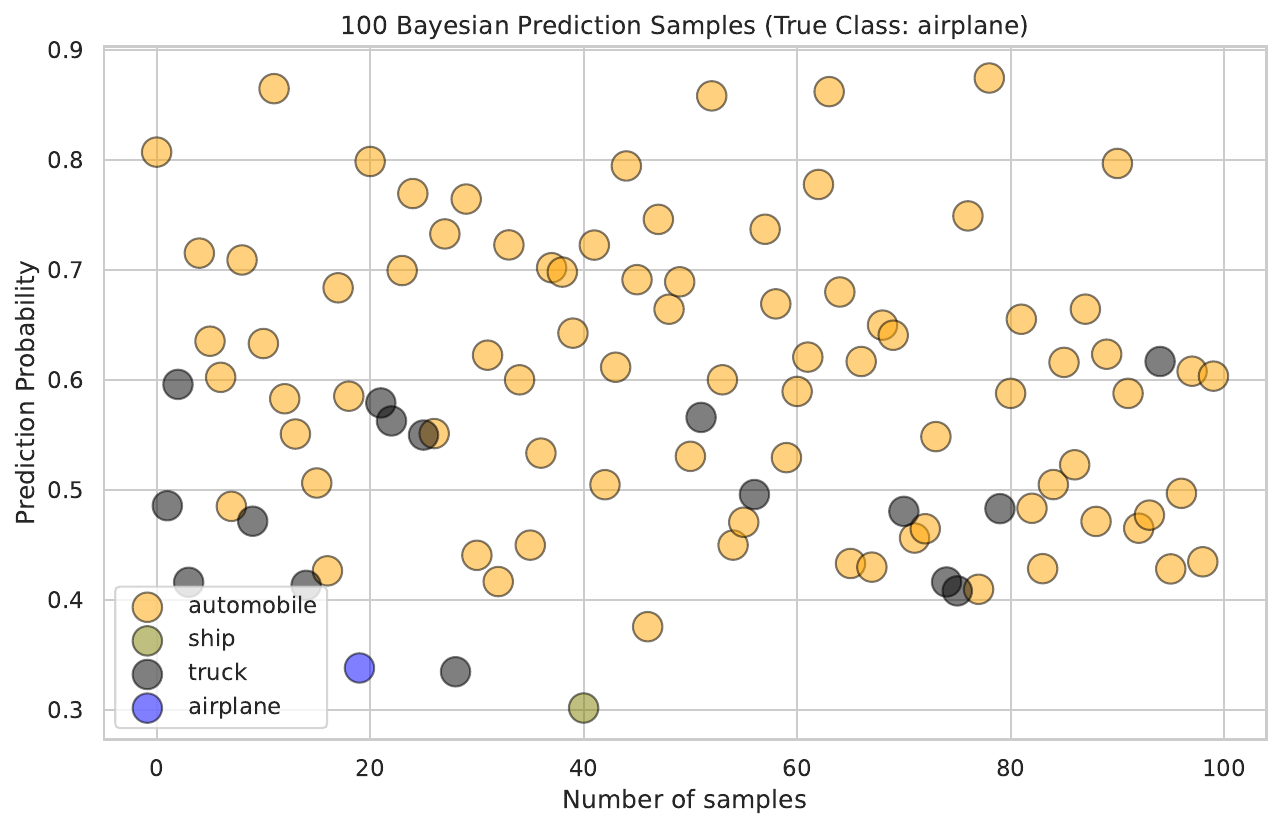}
    \end{minipage}
    \begin{minipage}[t]{0.49\textwidth}
        \centering
        \includegraphics[width=\textwidth]{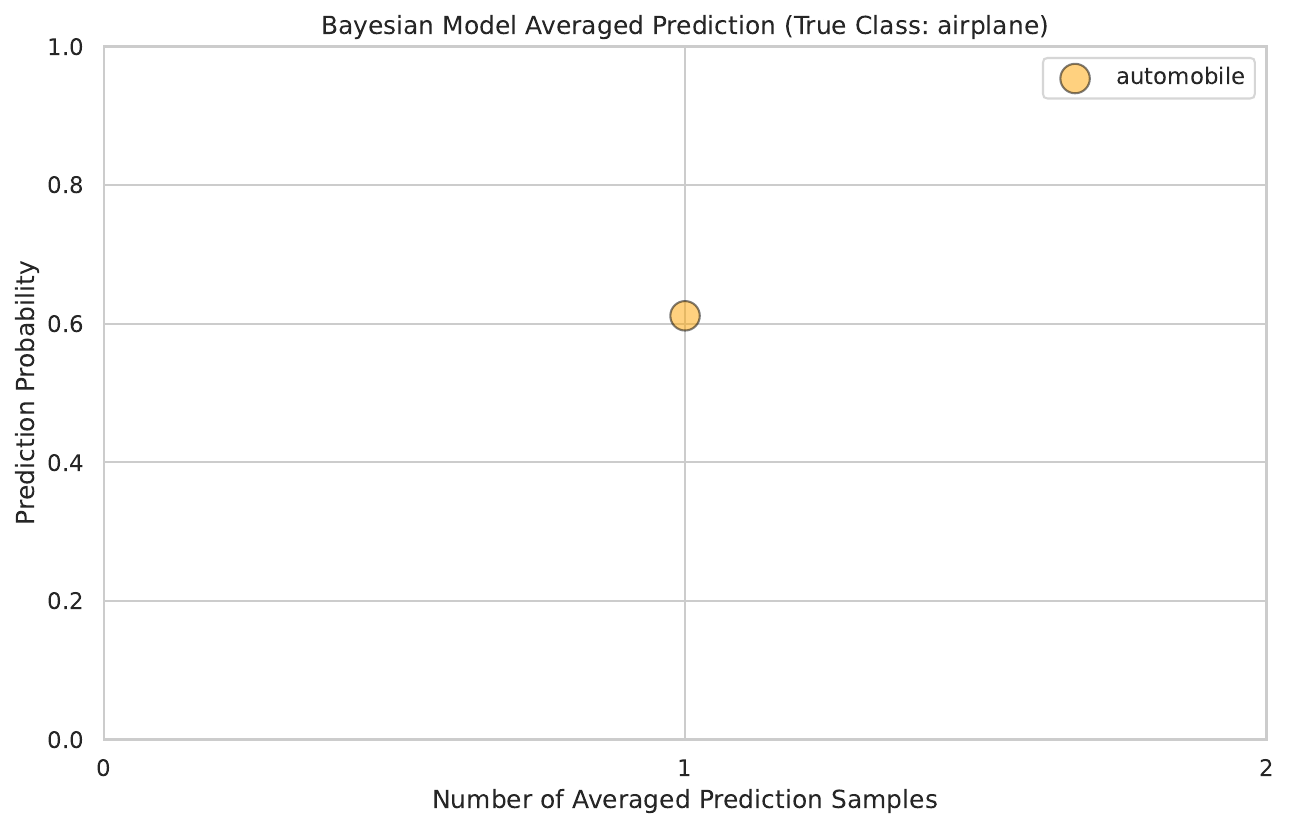}
    \end{minipage}\hfill
    \begin{minipage}[t]{0.49\textwidth}
        \centering
        \includegraphics[width=\textwidth]{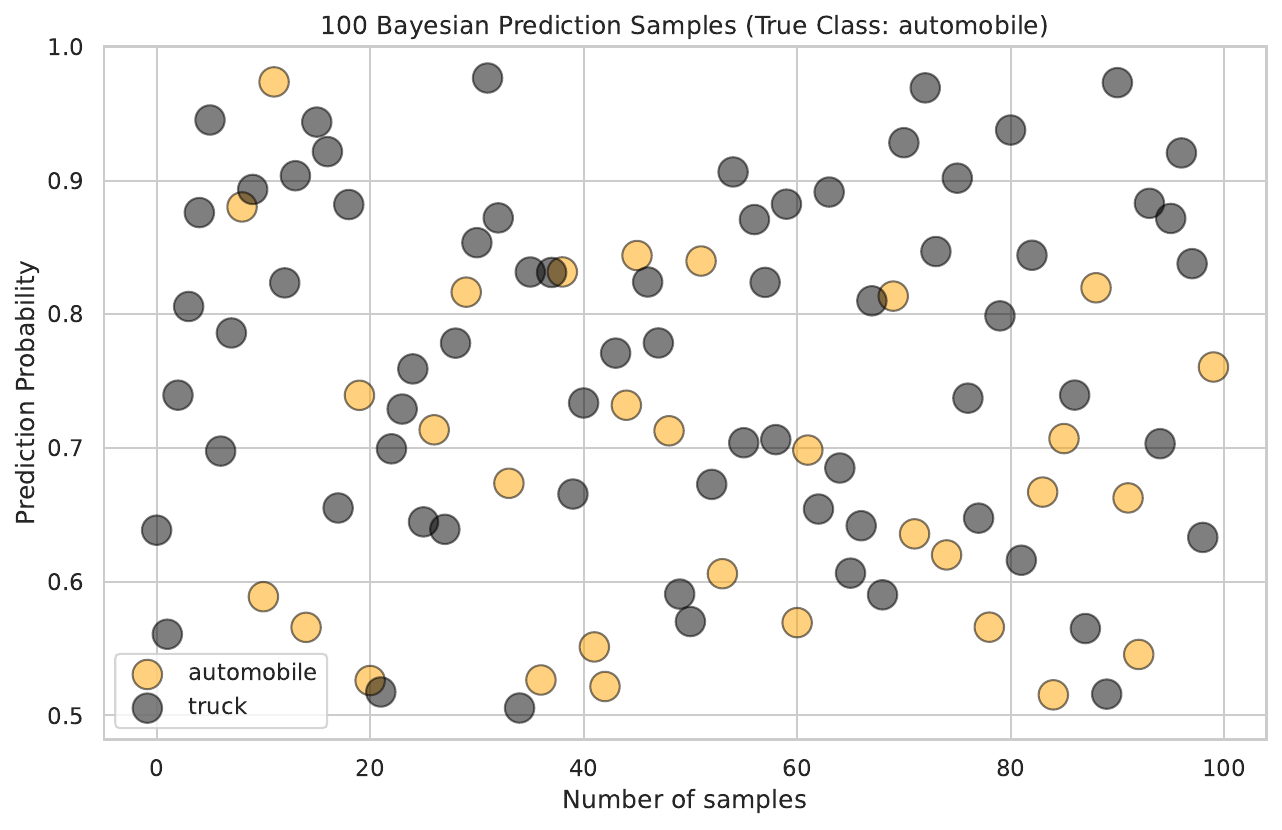}
    \end{minipage}
    \begin{minipage}[t]{0.49\textwidth}
        \centering
        \includegraphics[width=\textwidth]{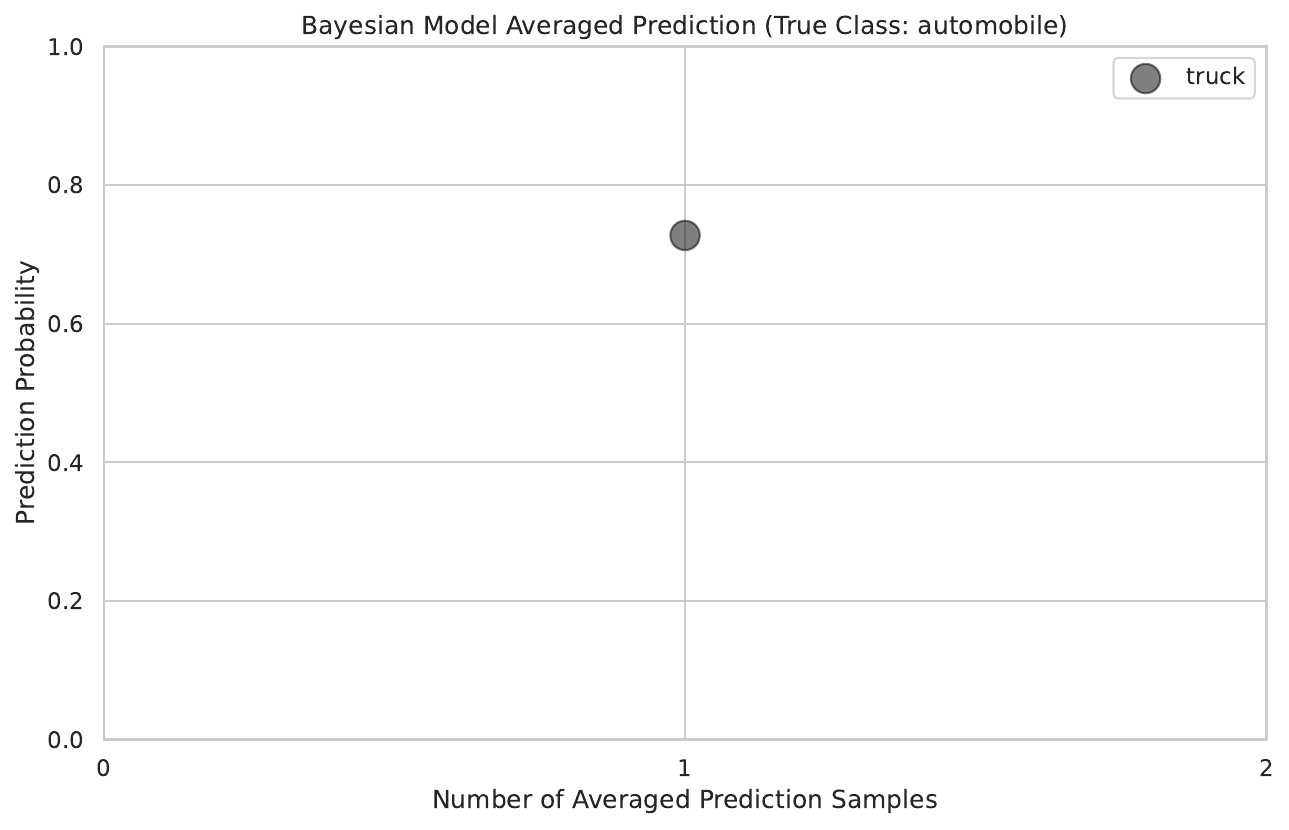}
    \end{minipage}
    \caption{Visualizations of 100 prediction samples obtained prior to Bayesian Model Averaging and corresponding Bayesian Model Averaged prediction in two real scenarios from CIFAR-10.}
    \label{fig:BMA_vs_nonBMA}
\end{figure}

In BNNs,
\emph{aleatoric uncertainty} is measured by the predictive entropy, while \emph{epistemic uncertainty} is represented by \emph{mutual information} \citep{hullermeier2021aleatoric}, MI, which measures the difference between the entropy of the predictive distribution and the expected entropy of the individual predictions.
To compute both mutual information 
and predictive entropy 
in Bayesian Neural Networks (BNNs), one utilises the predictive distributions of the model. 
MI quantifies the amount of information gained about the label $y$ given the input $\mathbf{x}$ and the observed data $\mathbb{D}$, while the predictive entropy ($H$) measures the uncertainty associated with the predictions:
\begin{equation}
    \text{MI}(\hat{p}_{b}(y \mid \mathbf{x}, \mathbb{D})) = H(\hat{p}_{b}(y \mid \mathbf{x}, \mathbb{D})) - \mathbb{E}_{\mathbb{D}}[H(p(y \mid \mathbf{x},{\theta}))] ,
\end{equation}
where $H(\hat{p}_{b}(y \mid \mathbf{x}, \mathbb{D}))$ is the entropy of the predictive distribution obtained from BMA, and $\mathbb{E}_{\mathbb{D}}[H(p(y \mid \mathbf{x},{\theta}))]$ represents the expected entropy of the individual predictive distributions sampled from the posterior distribution of the parameters $p(\theta \mid \mathbb{D})$. $H(\cdot)$ denotes the Shannon entropy function. 

The predictive entropy can be calculated as: 
\begin{equation}
H(\hat{p}_{b}(y \mid \mathbf{x}, \mathbb{D})) = - \int \hat{p}_{b}(y \mid \mathbf{x}, \mathbb{D}) \log \hat{p}_{b}(y \mid \mathbf{x}, \mathbb{D}) dy ,
\end{equation}
where $\hat{p}(y \mid \mathbf{x}, \mathbb{D})$ is the predictive distribution. This equation represents the average uncertainty associated with the predictions across different possible values of $y$, considering the variability introduced by the parameter uncertainty captured in the posterior distribution $p(\theta \mid \mathbb{D})$.

\vspace{-2pt}
\subsection{Deep Ensembles} 

In Deep Ensembles (DEs) \citep{lakshminarayanan2017simple}, 
a prediction 
$\hat{p}_{de}(y \mid \mathbf{x}, \mathbb{D})$ for an input $\mathbf{x}$ is obtained by averaging the predictions of $K$ individual models:
$
   \hat{p}_{de}(y \mid \mathbf{x}, \mathbb{D}) = \frac{1}{K} \sum_{k=1}^{K} \hat{p}_k(y \mid \mathbf{x}, \mathbb{D}) ,
$
where $\hat{p}_{k}$ represents the prediction of the $k$-th model, trained independently with different initialisations or architectures.

In Deep Ensembles,
{aleatoric uncertainty} is assessed via the predictive entropy, averaged entropy of each ensemble's prediction, while {epistemic uncertainty} is encoded by the predictive variance, 
the difference between the entropy of all ensembles and the averaged entropy of each ensemble.

Let $\mathcal{M} = \{ M_1, M_2, \ldots, M_K \}$ denote the ensemble of $K$ neural network models for $k = 1, 2, \ldots, K$. Given an input $\mathbf{x}$, the prediction $y_{\mathcal{M}}$ is obtained by averaging the predictions of individual models.
The \emph{predictive entropy} 
represents the averaged entropy of each ensemble's prediction $y_{k}$ given the input 
$\mathbf{x}$ and the observed data $\mathbb{D}$:
\begin{equation}
H(\hat{p}_{de}(y \mid \mathbf{x}, \mathbb{D})) = \frac{1}{K} \sum_{k=1}^{K} H(\hat{p}_{de}(y_{k} \mid \mathbf{x}, \mathbb{D})),
\end{equation}
where $y_{k}$ represents the prediction of the $k$-th model $M_k$.

The \emph{predictive variance} 
is measured as the difference between the entropy of all the ensembles, $H(\hat{p}_{de}(y_{\mathcal{M}} \mid \mathbf{x}, \mathbb{D}))$, and the averaged entropy of each ensemble, $H(\hat{p}_{de}(y \mid \mathbf{x}, \mathbb{D}))$.
\begin{equation}
H(y_{\mathcal{M}}) = H(\hat{p}_{de}(y_{\mathcal{M}} \mid \mathbf{x}, \mathbb{D})) - H(\hat{p}_{de}(y \mid \mathbf{x}, \mathbb{D})).
\end{equation}

The predictive variance in DEs is considered an approximation of mutual information \citep{hullermeier2021aleatoric}. This formulation captures both the model uncertainty inherent in the ensemble predictions and the uncertainty due to the variance among individual model predictions. While DEs have proven as a good baseline method for uncertainty quantification in practice, they remain computationally expensive with several recent methods aiming to approximate ensemble uncertainties with single models \citep{van2020uncertainty, lahlou2021deup, garipov2018loss, zanger2025contextual}. 

\subsection{Evidential Deep Learning} 

Evidential Deep Learning (EDL) models \citep{sensoy} make predictions $\hat{p}_{e}(y \mid \mathbf{x}, \mathbb{D})$ as parameters of a second-order Dirichlet distribution on the class space, instead of softmax probabilities. EDL uses these parameters to obtain a pointwise 
prediction.
Similar to BNNs, averaged DE and EDL predictions are point-wise predictions and averaging may not always be optimal.

\subsection{Deep Deterministic Uncertainty} 

Deep Deterministic Uncertainty (DDU) \citep{mukhoti2023deep} models 
differ from other uncertainty-aware baselines as they do not represent uncertainty in the prediction space, but do so in the input space by identifying whether an input sample is in-distribution (iD) or out-of-distribution (OoD). As a result, DDU provides predictions $\hat{p}_{ddu}(y \mid \mathbf{x}, \mathbb{D})$ in the form of softmax probabilities akin to traditional neural networks (NNs).

\subsection{Credal Models}

Models that generate \emph{credal sets} \citep{levi80book,zaffalon-treebased,cuzzolin2010credal,antonucci2010credal,cuzzolin2008credal} 
represent uncertainty in predictions by providing a set of plausible outcomes, rather than a single point estimate. 
A \emph{credal set} \citep{levi80book,zaffalon-treebased,cuzzolin2010credal,antonucci2010credal,cuzzolin2008credal} is a convex set of probability distributions on the target (class) space. 
Credal sets can be elicited, for instance, 
from predicted
probability intervals \citep{wang2025creinns, caprio2023imprecise} 
$[\hat{\underline{p}}(y), \hat{\overline{p}}(y)]$, encoding lower and upper bounds, respectively, to the probabilities of each of the classes:
\begin{equation}\label{eq:credal_interval}
    \hat{\mathbb{C}r} (y \mid \mathbf{x}, \mathbb{D}) = \{ p \in \mathcal{P} \ | \ \hat{\underline{p}}(y) \leq p(y) \leq \hat{\overline{p}}(y), \forall y \in \mathbf{Y} \}.
\end{equation}
A credal set is efficiently represented by its extremal points; their number can vary, depending on the size of the class set and the complexity of the network prediction the credal set represents.

\subsection{Belief Function Models} 
\label{sec:belief-functions}

\emph{Belief functions} \citep{Shafer76} 
are non-additive measures independently assigning a degree of belief to each subset $A$ of their sample space, 
indicating the support for that subset. 
\\
A predicted belief function $\hat{Bel}$ on 
$\mathbf{Y}$ is mathematically equivalent to the credal set 
\vspace{-2mm}
\begin{equation} \label{eq:consistent}
\mathbb{C}r_{\hat{Bel}} (y \mid \mathbf{x}, \mathbb{D}) 
=
\big \{ 
p \in \mathcal{P}  \; \big | \; p(A) \geq \hat{Bel}(A) 
\big \}.
\vspace{-1mm}
\end{equation} 
Its center of mass, termed 
\emph{pignistic probability} \citep{SMETS2005133} $BetP[\hat{Bel}]$,
assumes the role of the predictive distribution for belief function models
 \citep{tong2021evidential, manchingal2025randomsetneuralnetworksrsnn}:
$
   \hat{p}_{bel}(y \mid \mathbf{x}, \mathbb{D}) = 
    BetP[\hat{Bel}]. 
$

Belief functions can be derived from \emph{mass functions} through a normalization process, where the belief assigned to a hypothesis is the sum of the masses of all subsets of the frame of discernment that include the hypothesis.
A {mass function} 
\citep{Shafer76} is a set function \citep{denneberg99interaction} $m : 2^\Theta\rightarrow[0,1]$ such that $m(\emptyset)=0$ and
$\sum_{A\subset\Theta} m(A)=1$. In classification, $2^\Theta$ is the set of all subsets of classes $\mathcal{C}$, 
the powerset $\mathbb{P}(\mathcal{C})$. 
Subsets of $\Theta = \mathcal{C}$ whose mass values are non-zero are called \emph{focal elements} of $m$.
The \emph{belief function} associated with 
$m$ 
is given by:
$
Bel(A) = \sum_{B\subseteq A} m(B). 
$
The redistribution of mass values back to singletons from focal sets is achieved through the concept of \emph{pignistic probability} \citep{SMETS2005133}. Pignistic probability ($BetP$), also known as Smets' pignistic transform, is a method used to assign precise probability values to individual events based on the belief function's output. 

\textit{Aleatoric uncertainty} in such models is represented as the pignistic entropy of predictions $H_{BetP}$, whereas \textit{epistemic uncertainty} can be modelled by the `size' of the credal set (Eq. \ref{eq:consistent}). 


\section{A comparison of uncertainty estimation models}
\label{app:comparison-experiments}

In Tab. \ref{tab:uq-comparison-table}, we present the training and inference times (computational costs) for the uncertainty methods discussed in Sec. \ref{sec:uncertainty-models} and Fig. \ref{fig:uncertainty-models}. Two examples of each model type are shown, all trained on the ResNet50 backbone. More training details are given below.

The models evaluated include a range of uncertainty estimation frameworks: traditional model (ResNet50), Bayesian approximations such as Laplace \citep{hobbhahn2022fast} and function-space variational inference \citep{rudner2022tractable}, ensemble methods including deep ensembles \citep{lakshminarayanan2017simple} and epistemic neural networks (ENN) \citep{osband2024epistemic}, evidential approaches \citep{sensoy, qu2024hyper}, and Epistemic AI frameworks based on credal sets \citep{wang2024credal} and random-set theory \citep{manchingal2025randomsetneuralnetworksrsnn}. These models differ not only in their uncertainty modeling principles but also in computational costs (see Tab. \ref{tab:uq-comparison-table}), reflecting a spectrum of trade-offs between performance and efficiency. For instance, function-space Bayesian methods provide uncertainty but at a high computational cost \citep{rudner2022tractable}, while epistemic random-set models offer competitive accuracy with efficient inference \citep{manchingal2025randomsetneuralnetworksrsnn}.


\begin{table}[!ht]
\caption{Training (in minutes; per 100 epochs) and inference time (in milliseconds; per sample) comparison of uncertainty estimation methods on the CIFAR-10 dataset.}
\label{tab:uq-comparison-table}
\vspace{-5pt}
\begin{center}
\resizebox{\textwidth}{!}{%
\begin{sc}
\begin{tabular}{lccc}
\toprule
\textbf{Model}         & \textbf{Training Time (100 epochs) (min)} & \textbf{Inference Time (ms/sample)} \\
\midrule
Traditional                      & 85.33                  & 1.91 $\pm$ 0.7                                            \\
Deterministic (DDU) \citep{mukhoti2023deep} &  243.85    &   59.35   $\pm$ 0.40   \\
Bayesian (Laplace) \citep{hobbhahn2022fast}  & 107.90                          & 7.11 $\pm$ 0.89                                            \\
Bayesian (Function SVI) \citep{rudner2022tractable}   & 1518.35                         & 340.25 $\pm$ 0.76                                               \\
Ensemble (Deep Ensembles) \citep{lakshminarayanan2017simple}  & 426.66                     & 13163.50 $\pm$ 3.37                                             \\
Ensemble (ENN) \citep{osband2024epistemic}      & 712.30                          & 3.10 $\pm$ 0.03                                                 \\
Evidential (EDL) \citep{sensoy} & 188.57 & 6.12 $\pm$ 0.01\\
Evidential (Hyper-Opinion EDL) \citep{qu2024hyper} & 186.56  & 23.01 $\pm$ 0.15\\
Epistemic AI (Credal) \citep{wang2024credal}                  & 122.95                   & 63.0 $\pm$ 1.1                                          \\
Epistemic AI (Random-Set) \citep{manchingal2025randomsetneuralnetworksrsnn}         & 113.23                                   & 1.91 $\pm$ 0.02                      \\
\bottomrule
\end{tabular}
\end{sc}
}
\end{center}
\end{table}

\textbf{Training details.} All models were trained using a ResNet50 backbone (excluding the final classification layer), followed by two additional dense layers with 1024 and 512 neurons, respectively, using ReLU activation. For the \textit{Epistemic: Random-set} \citep{manchingal2025randomsetneuralnetworksrsnn} model, the output layer used a sigmoid activation function to support multi-label classification, while all other models, \textit{Epistemic: Interval} \citep{wang2025creinns}, \textit{Epistemic: Credal} \citep{wang2024credal}, \textit{Epistemic: Wrapper} \citep{wang2024credalwrapper}, \textit{Bayesian: Laplace} \citep{hobbhahn2022fast}, \textit{Bayesian: Function SVI} \citep{rudner2022tractable}, \textit{Ensemble: Deep} \citep{lakshminarayanan2017simple}, and \textit{Ensemble: ENN} \citep{osband2024epistemic}, used a softmax output for multi-class classification. The initial learning rate was set to 1e-3, with a scheduler that reduced the rate by a factor of 0.1 at epochs 80, 120, 160, and 180. Models were trained for 200 epochs using a batch size of 128. The optimizer varied by model: Adam was used for \textit{Epistemic: Random-set}, \textit{Epistemic: Wrapper}, \textit{Ensemble: ENN}, and \textit{Ensemble: Deep}, while \textit{Bayesian: Function SVI} used SGD. Training \textbf{dataset} sizes were as follows: CIFAR-10 used 40,000 samples and ImageNet used 1,172,498 samples. Test datasets contained 10,000 samples for CIFAR-10 and 2,000 for ImageNet. For out-of-distribution (OoD) evaluation, 10,000 test samples were used. All models were trained and evaluated using 224×224 input image size with data augmentation including random horizontal/vertical shifts (magnitude 0.1) and horizontal flips. Experiments were conducted using an NVIDIA A100 80GB GPU.

The pairwise plots in Fig. \ref{fig:pairwise-plots} illustrate the relationships between key uncertainty and performance metrics: Entropy, Expected Calibration Error (ECE), Area Under the Receiver Operating Characteristic curve (AUROC), and Area Under the Precision-Recall Curve (AUPRC), across different uncertainty estimation methods. These methods are categorized into Epistemic AI and competitor types, revealing distinct clusters and trends. Notably, Epistemic AI models show higher entropy values and competitive AUROC/AUPRC scores, indicating richer uncertainty quantification alongside robust out-of-distribution (OoD) detection. In contrast, the competitor generally exhibit lower entropy and slightly varied calibration performance. The correlations visible in the plots reflect inherent trade-offs: higher uncertainty often aligns with better OoD detection but may impact calibration, which is crucial for reliable decision-making.

\begin{figure}[!h]
    \centering
    \includegraphics[width=\textwidth]{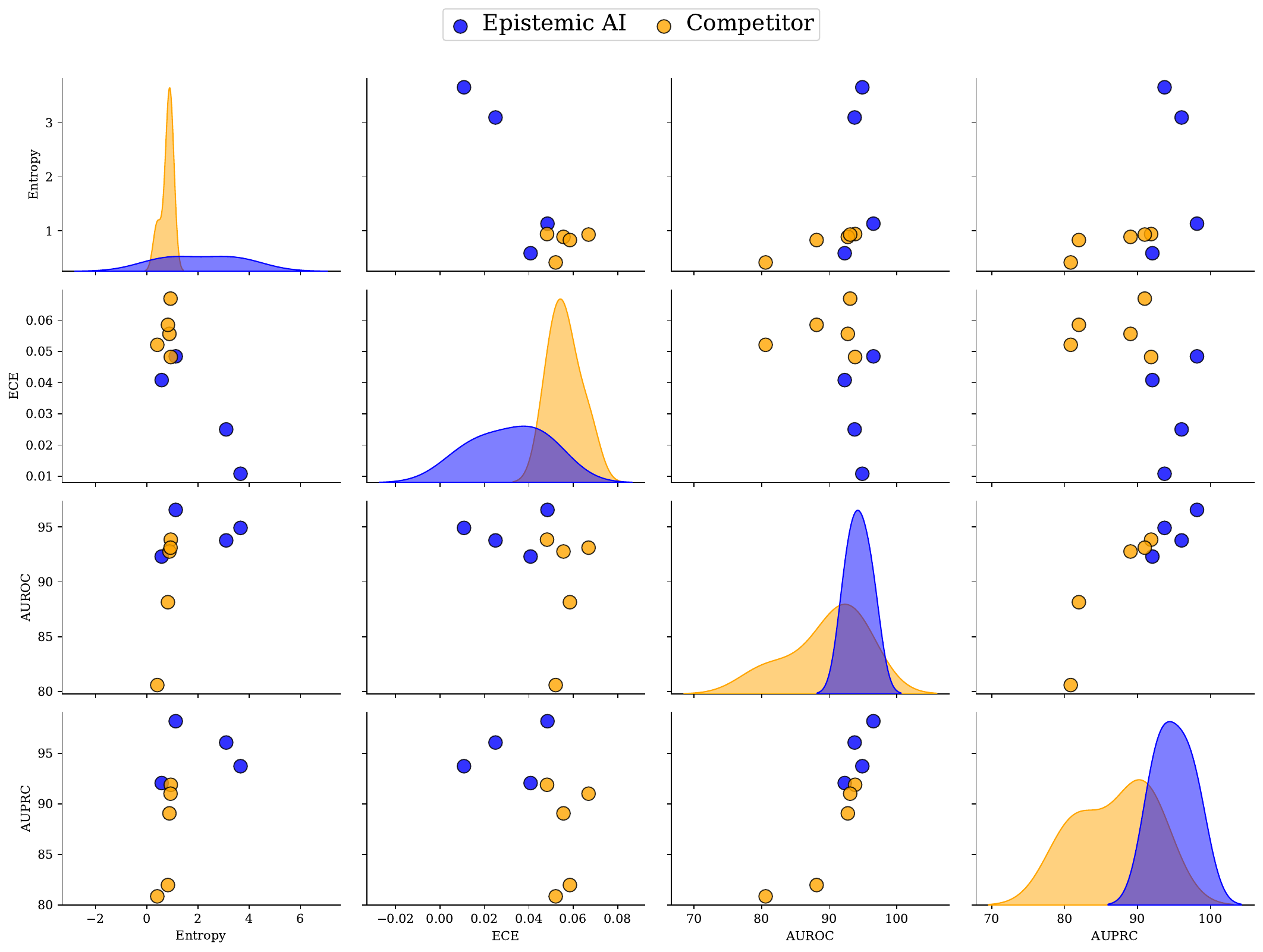}
    \caption{Pairwise comparison of uncertainty and performance metrics for various uncertainty estimation methods on CIFAR-10. Metrics include Entropy, Expected Calibration Error (ECE), AUROC, and AUPRC (OoD Detection).
}
    \label{fig:pairwise-plots}
\end{figure}

\vspace{-6pt}
\section*{Impact Statement}

This work advances Epistemic AI, offering a more reliable and interpretable approach to uncertainty quantification in machine learning. By improving AI's ability to distinguish between known and unknown uncertainties, this research enhances robustness in critical applications such as healthcare, autonomous systems, climate modeling, and scientific discovery. A key ethical advantage is its potential to mitigate overconfidence in AI predictions, reducing risks in safety-critical domains like medical diagnosis and autonomous decision-making.

Future societal impacts include more trustworthy AI systems that can adapt to novel and evolving situations, fostering responsible deployment in high-stakes environments. Furthermore, integrating epistemic uncertainty into AI could bridge gaps between symbolic reasoning and deep learning, advancing neurosymbolic AI and promoting generalizable, human-aligned decision-making. However, ethical considerations only arise in the potential misuse of uncertainty-aware AI, such as adversarial exploitation or biased decision-making if epistemic uncertainty is misinterpreted. Addressing these risks requires transparent AI models, regulatory oversight, and interdisciplinary collaboration.

\end{document}